\documentclass{article}

\usepackage{amssymb}
\usepackage{amsmath}
\usepackage[dvipsnames,table]{xcolor}   %
\usepackage{colortbl}      
\usepackage{wrapfig}
\usepackage{subcaption} %
\usepackage{graphicx}
\usepackage{booktabs}
\usepackage{caption}   %
\usepackage{eucal}
\usepackage{tikz}
\newif\ifdrafting
\draftingtrue %
\ifdrafting
    \newcommand{\todo}[1]{{\leavevmode\color[rgb]{1,0,0}[TODO: #1]}}
    \newcommand{\jz}[1]{{\leavevmode\color[rgb]{0,0.8,0.2}[Junyi: #1]}}
    \newcommand{\ak}[1]{{\leavevmode\color[rgb]{0,0.6,0.8}[AK: #1]}}
    \newcommand{\TD}[1]{{\leavevmode\color[rgb]{0,0.8,0.9}[TD: #1]}}
    \newcommand{\hs}[1]{{\leavevmode\color[rgb]{0.8,0.2,0.8}[Hongsuk: #1]}}
    
    \newcommand{\arthur}[1]{{\leavevmode\color[rgb]{1.0,0.0,0.0}[Arthur: #1]}}
\else
    \newcommand{\todo}[1]{}
    \newcommand{\jz}[1]{}
    \newcommand{\ak}[1]{}
    \newcommand{\TD}[1]{}
    \newcommand{\hs}[1]{}
    \newcommand{\arthur}[1]{}
\fi

\usepackage{bbm}
\newcommand{\cmark}{{\color{green!60!black}\ding{51}}}
\newcommand{\xmark}{{\color{red}\ding{55}}}

\usepackage[final]{corl_2025}
\usepackage{pifont}

\title{{Visual Imitation Enables Contextual Humanoid Control}}

\title{
\makebox[\textwidth]{{
\hspace{1pt}
Visual Imitation Enables Contextual Humanoid Control
}
}
}
\author{%
\\
\vspace{-2.5em}
\\
  \hspace{-20pt}
  \normalfont{Arthur Allshire}\footnotemark[1]
  \quad
  Hongsuk Choi\footnotemark[1]%
  \quad
  Junyi Zhang\footnotemark[1]%
  \quad
  David McAllister\footnotemark[1]%
  \quad 
  Anthony Zhang%
  \quad
  Chung Min Kim% 
  \\ [0.2em]
  Trevor Darrell%
  \quad
  Pieter Abbeel%
  \quad 
  Jitendra Malik%
  \quad
  Angjoo Kanazawa
  \\
  \quad \vspace{-0.5em}
  \\
    \hspace{-20pt}
    \textbf{UC Berkeley} \\
    \quad \vspace{-0.2em}
  \\
}

\date{}  % no date

\begin{document}
\maketitle

{\renewcommand{\thefootnote}{}\footnotetext{\scalebox{0.95}{\noindent\hspace{-1em} $^*$These authors contributed equally. Correspondence to: \texttt{allshire@berkeley.edu}, \texttt{hongsuk@berkeley.edu}.}}}

% now put your “teaser” figure *after* the title rather than inside \@maketitle
\vspace{-3em}
\begin{figure}[htbp]
  \centering
  \includegraphics[width=\linewidth]{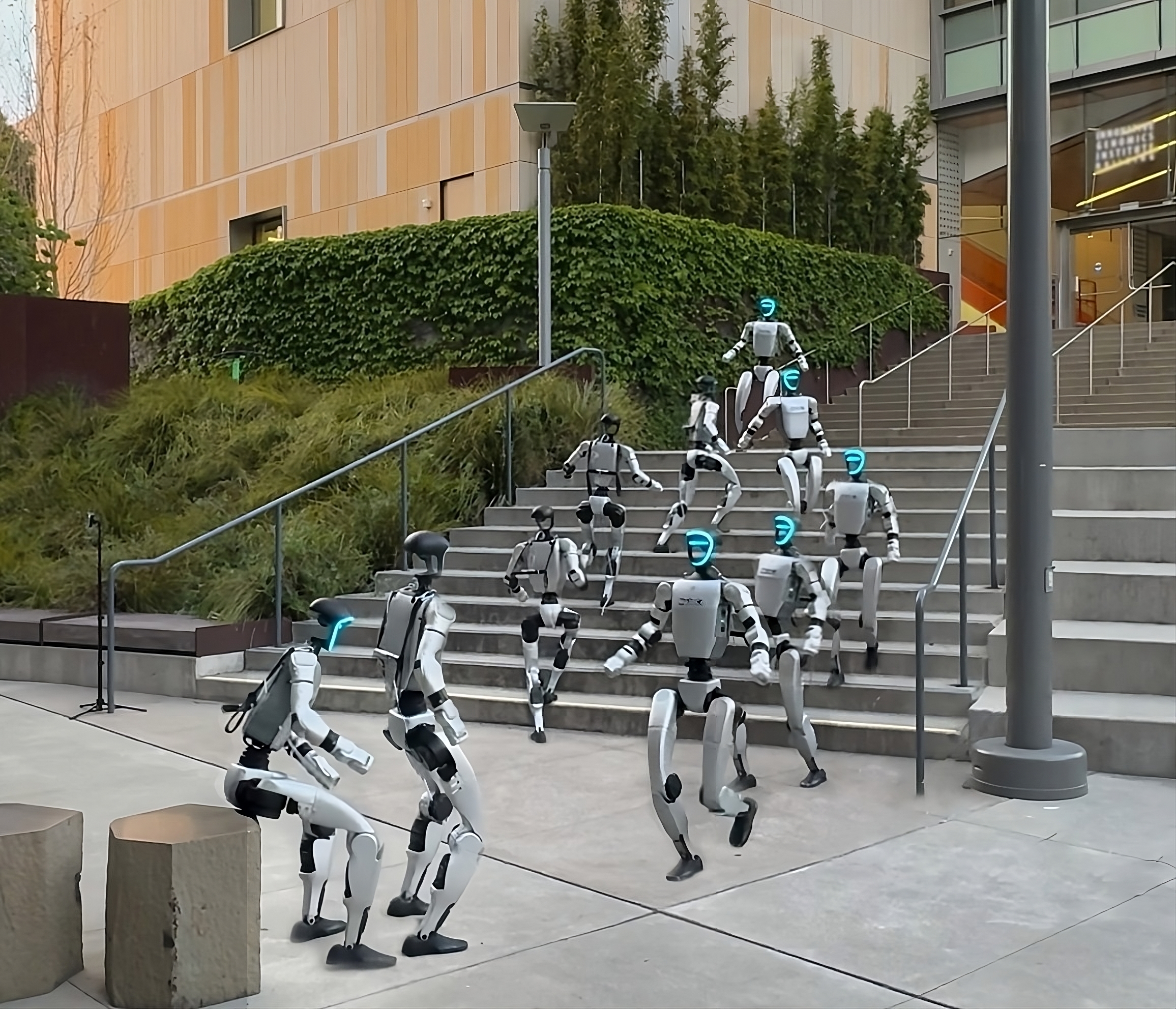}
  \caption{\small \textbf{\textsc{VideoMimic}} is a real-to-sim-to-real pipeline that converts monocular videos into transferable humanoid skills, letting robots learn context-aware behaviors (terrain-traversing, stairs-climbing, sitting) in a single policy. Video results are available on our webpage: \url{https://videomimic.net}.}
  \label{fig:teaser}
\end{figure}
\vspace{-1em}

\begin{abstract}
{
     How can we teach humanoids to climb staircases and sit on chairs using the surrounding environment context? Arguably, the simplest way is to \emph{just show them}---casually capture a human motion video and feed it to humanoids.
     We introduce \textsc{VideoMimic}, a real-to-sim-to-real pipeline that mines everyday videos, jointly reconstructs the humans and the environment, and produces whole-body control policies for humanoid robots that perform the corresponding skills.
    We demonstrate the results of our pipeline on real humanoid robots, showing robust, repeatable contextual control such as staircase ascents and descents, sitting and standing from chairs and benches, as well as other dynamic whole-body skills---all from a single policy, conditioned on the environment and global root commands. %
    \textsc{VideoMimic} offers a scalable path towards teaching humanoids to operate in diverse real-world environments.
    }
     \end{abstract}

% \keywords{Visual Imitation, Humanoids, Reinforcement Learning, 4D Reconstruction from an RGB video} 

\section{Introduction}
\label{sec:introduction}

How do we learn to interact with the world around us---like sitting on a chair or climbing a staircase? We watch others perform these actions, try them ourselves, and gradually build up the skill. Over time, we can handle new chairs and staircases, even if we have not seen those exact ones before. If humanoid robots could learn in this way---by observing everyday videos---they could acquire diverse contextual whole-body skills without relying on hand-tuned rewards or motion-capture data for each new behavior and environment. We refer to this ability to execute environment-appropriate actions as contextual control.

We introduce \textsc{VideoMimic}, a real-to-sim-to-real pipeline that turns monocular videos—such as casual smartphone captures—into transferable skills for humanoids. From these videos, we jointly recover the 4D human-scene geometry, retarget the motion to a humanoid, and train an RL policy to track the reference trajectories.
We then distill the policy %
into a single unified policy that observes only proprioception, a local height-map, and the desired root direction. This distilled policy outputs low-level motor actions conditioned on the terrain and body state, allowing it to execute appropriate behaviors—such as stepping, climbing, or sitting—across unseen environments without explicit task labels or skill selection.

We develop a perception module that reconstructs 3D human motion %
from a monocular RGB video, along with aligned scene point clouds in the world coordinate frame.
We convert the point clouds into meshes and align them with gravity to ensure compatibility with physics simulators.
The global motion and local poses are retargeted to a humanoid with constraints that ensure physical plausibility, accounting for the embodiment gap.
The mesh and retargeted data seed a goal-conditioned DeepMimic~\cite{DeepMimic}-style reinforcement-learning phase in simulation: we warm-start on MoCap data, then train a single policy to track motions from multiple videos in their respective height-mapped environments 
while randomizing mass, friction, latency, and sensor noise for robustness. 
Once our tracking policy is trained, we distill it using DAgger \cite{dagger} to a policy %
that operates without conditioning on target joint angles.
The new policy observes proprioception, an 11 × 11 height-map patch centered on the torso, and the vector to the goal in the robot's local reference frame. 
PPO fine-tuning under this reduced observation set yields a generalist controller that, given height-map and root direction at test time, selects and smoothly executes context-appropriate actions such as stepping, climbing, or sitting. %
In particular, every step of our policy relies only on observations available at real-world deployment, making it immediately runnable on real hardware. %

Our approach bridges 4D video reconstruction and robot skill learning in a single, data-driven loop. Unlike earlier work that recovers only the person or the scene in isolation, we jointly reconstruct both at a physically meaningful scale and represent them as meshes and motion trajectories suitable for physics-based policy learning.
We train our approach on 123 monocular RGB videos, which will be released. 
We validate the approach through deployment on a real Unitree G1 robot, which shows generalized humanoid motor skills in the context of surrounding environments, even on unseen environments.
We will release the reconstruction code, policy training framework, and the video dataset to facilitate future research.

\section{Related Work}
\label{sec:related_work}

\noindent\textbf{Learning Skills on Legged Robots.}

\vspace{-0.5em}
\begin{wraptable}[8]{r}{0.58\textwidth}  %
  \centering
  \vspace{-4.75em}
  \resizebox{\linewidth}{!}{%
    \begin{tabular}{lccc}
      \toprule
      \textbf{Method} & \textbf{Env. Real-to-Sim} & \textbf{Context. Ctrl} & \textbf{Real Robot} \\
      \midrule
      \rowcolor{gray!10}
      DeepMimic / SfV \cite{DeepMimic,SfV}           & \xmark & \xmark & \xmark \\
      Egocentric Loco \cite{egoloco}       & \xmark & \cmark & \cmark \\
      \rowcolor{gray!10}
      ASAP \cite{ASAP}                     & \xmark & \xmark & \cmark \\
      Humanoid Loco. \cite{radosavovic2023} & \xmark & \xmark & \cmark \\
      \rowcolor{gray!10}
      H2O / ExBody2 \cite{h2o,exbody2}         & \xmark & \xmark & \cmark \\
      Parkour \cite{anymalparkour, humanoidparkourlearning} & \xmark & \cmark & \cmark \\
      \rowcolor{gray!10}
      \midrule
      \textbf{VideoMimic (Ours)}                       & \cmark & \cmark & \cmark \\
      \bottomrule
    \end{tabular}
  }
  \caption{\textbf{Comparison of methods across different features.} \textsc{VideoMimic} transfers both human motion and scene geometry from real videos to simulation, learns context-aware control in simulation, and successfully deploys the resulting policy on real-world environments.}
  \label{table:checkmark_comparisons}
\end{wraptable}
Recent progress in legged-robot motor skills follows two complementary streams. Reward-based methods use model-free RL in simulation, shaping behavior with handcrafted objectives that mix task terms (e.g., velocity tracking) and motion-naturalness regularizers; thanks to massive parallel physics engines~\cite{raisim, isaacgym}, this paradigm has produced agile locomotion on quadrupeds and humanoids without motion data. However, each new behavior demands tuning of user-defined rewards and environment scripting~\cite{hwangbo19,joonholee2020quadruped,RMA,egoloco,radosavovic2023,anymalparkour, wang2025beamdojo, long2024learning}. Data-driven methods instead imitate reference motion, originally MoCap clips or monocular video, training a simulated character to track them and porting the idea to robots \cite{DeepMimic,SfV,yu2021human,luo2024universal,h2o,exbody2}. 
For example, recent work~\cite{radosavovic2024learning,radosavovic2024humanoid} frames legged locomotion as a next-
token prediction task and pre-trains a policy on human data in kinematic space, showing strong performance.
While imitation bypasses reward engineering~\cite{ASAP}, existing works typically assume flat ground or manually designed setups, limiting context-aware whole-body control; even animation systems that model human-scene contact rely on instrumented MoCap stages and thus lack scalability \cite{hassan2023synthesizing,tokenhsi}. 
Our system conditions on visual observations, the local height‑maps, and learns environment‑aware skills such as stair‑climbing and chair‑sitting directly from monocular RGB videos. Joint 4D human–scene reconstruction provides physically consistent reference motions, which RL distills into policies that transfer to a real humanoid (Table~\ref{table:checkmark_comparisons}).

\noindent\textbf{Human and Scene Reconstruction from Images and Videos.}  
Early monocular-video methods regress pose and shape of humans~\cite{loper2023smpl} in a camera-relative frame with deep networks~\cite{loper2023smpl,kanazawa2018hmr,kocabas2020vibe}, which suffices for rendering, action recognition, or single-person tracking \citep{peng2021neural,moon2024expressive,rajasegaran2022tracking,luvizon20182d,rajasegaran2023benefits} but leaves the global trajectory---and thus context-aware dynamics---undefined; pioneers like SfV hand-tuned a global scale and even assumed a static camera, limiting generality. Recent methods combine human motion priors with SfM/SLAM to recover metric trajectories~\cite{ye2023slahmr,yuan2022glamr,kocabas2024pace}, yet still model only the person and camera. Advances in general scene parsing~\cite{li2024megasam,zhang2024monst3r,wang2025continuous} have enabled joint human-scene reconstruction that resolves scale via multi-view cues or learned priors~\citep{muller2024hsfm,liu2025josh}, but these systems have not been validated on robots. Parallel work injects physics constraints in post-processing or simulation~\cite{yuan2021simpoe,yuan2023tennis,ugrinovic2024multiphys,zhang2021lemo,li2022dnd}, trading scalability for realism. Our pipeline unifies these threads: it simultaneously estimates metric human motion and surrounding geometry from in-the-wild videos---without MoCap, pre-scanned scenes, or reward engineering---and outputs simulator-ready trajectories that respect contacts and collisions, enabling scalable learning of whole-body humanoid skills.

\begin{figure*}[t]
    \centering
    \makebox[\textwidth][c]{%
        \includegraphics[width=1.0\linewidth,clip,trim=0cm 0cm 0cm 0cm]{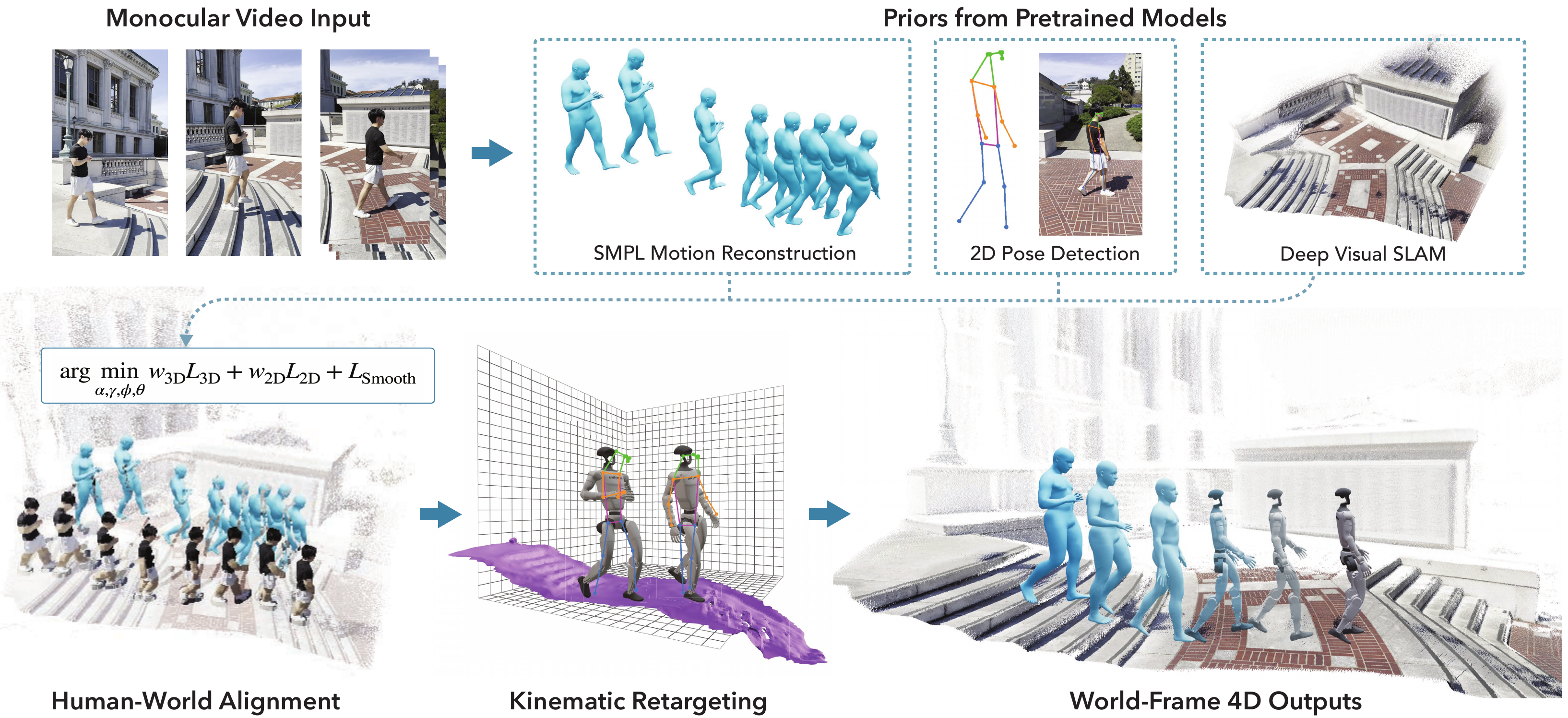}
    }
    \caption[]{\textbf{VideoMimic Real-to-Sim.} A casually captured phone video provides the only input. We first reconstruct per-frame human motion and 2D keypoints, along with a dense scene point cloud. An efficient optimization jointly aligns the motion and point cloud, recovers statistically accurate metric scale using a human height prior, and registers the human trajectory based on human-associated points. The point cloud is then converted to a mesh, aligned with gravity, and the motion is retargeted to a humanoid in the reconstructed scene. This yields world-frame trajectories and simulator-ready meshes that serve as inputs for policy training.
    }
    \label{fig:real2sim_method}
    \vspace{-1em}
\end{figure*}

\section{Real-to-Sim Data Acquisition}
Our real-to-sim pipeline proceeds as summarized in Figure~\ref{fig:real2sim_method}. We extract per-frame human poses and a raw scene point cloud from the input video (Sec.~\ref{subsec:reconstruction_preprocessing}); jointly optimize them to obtain metrically aligned human trajectories and scene geometry (Sec.~\ref{subsec:joint_reconstruction}); apply gravity alignment and convert the filtered point cloud into a lightweight mesh (Sec.~\ref{subsec:meshification}); and retarget the refined trajectories to the humanoid under joint-limit, contact, and collision constraints (Sec.~\ref{subsec:retargeting}). %
The resulting motion–mesh pairs are ready for policy learning in Sec.~\ref{sec:rl}.

\label{sec:real2sim}
\subsection{Preprocessing}  
\label{subsec:reconstruction_preprocessing}
We preprocess monocular RGB videos with off-the-shelf state-of-the-art human pose estimation and SfM methods.
First, people are detected and associated across frames using Grounded SAM2~\cite{ravi2024sam2,ren2024grounded}. For each detected person, we recover per‑frame 3D SMPL~\cite{loper2023smpl} parameters with VIMO~\cite{wang2024tram}, obtaining per‑frame local pose $\theta^{t}$, shape $\beta$, and SMPL 3D joints $J_{\mathrm{3D}}^{\,t} \in \mathbb{R}^{J\times3}$.
We detect 2D keypoints $ J_{\mathrm{2D}}^{\,t}$, i.e., body joint pixel positions, with ViTPose~\cite{xu2022vitpose}. 
Foot contact is regressed by BSTRO~\cite{huang2022rich}.
For scene reconstruction, we obtain the world point cloud from either MegaSaM~\cite{li2024megasam} or MonST3R~\cite{zhang2024monst3r}, which is parameterized as per-frame depth $D^t$, camera pose $[R^t|t^t]$, and a shared camera intrinsic matrix $K$.  
Note that the resulting point cloud is not metrically accurate.

To coarsely position the person in the world frame, we follow the initialization strategy of SLAHMR~\cite{ye2023slahmr}, using (i) the camera focal length predicted by SfM and (ii) the ratio between the average 2D limb length from the ViTPose detections $\tilde J_{\mathrm{2D}}^{\,t}$ and the corresponding metric scale 3D limb length in $J_{\mathrm{3D}}^{\,t}$, we estimate a similarity factor per frame that yields a coarse global trajectory $(\phi^{{t}_{0}}, \gamma^{{t}_{0}})$. Separately, we also lift $\tilde J_{\mathrm{2D}}^{\,t}$ to 3D by un‑projecting each pixel $(u,v)$ with its depth $D_{u,v}^{t}$ and intrinsics $K$ from SfM:
$\tilde J_{\mathrm{3D}}^{\,t} = K^{-1}[u,v,1]^{\top} D_{u,v}^{t}$. The lifted joints are then used to jointly optimize human poses and scene geometry scale, as described in the following section.

\subsection{Joint Human–Scene Reconstruction}
\label{subsec:joint_reconstruction}

Our pipeline jointly optimizes the human trajectory and the scene scale.  
The variables are the humans’ global translations $\gamma^{1:T}$, global orientations $\phi^{1:T}$, local poses $\theta^{1:T}$, and the scene point‑cloud scale $\alpha$.  
Because MegaSam pointclouds are scale‑ambiguous, the metric human height prior in the SMPL body models serves as the metric reference, while the lifted joints $\tilde J_{\mathrm{3D}}^{\,t}$ refine both the global trajectory and the local pose. We therefore solve for $\alpha$ simultaneously, reconciling any residual mismatch between the human‑derived scale and the scene geometry.

Inspired by He et al.~\cite{h2o}, we optionally run a scale-adaptation pass that searches for an SMPL shape $\beta^{\star}$ whose height and limb proportions match those of the G1 robot, prior to joint human-scene optimization.
The SMPL joints are then extracted from the reshaped mesh.
This use of a prefitted G1-scale SMPL $\beta$ effectively rescales the scene geometry to G1 size, improving the feasibility of humanoid motion—e.g., enabling actions like running or climbing over large obstacles—and facilitating reference motion learning. For real-world deployment, we skip this step and operate directly on the original metric-scale scene.

The objective combines joint-distance losses in 3D ($L_{3\text{D}}$), computed as the L1 distance between $\tilde J_{\mathrm{3D}}^{\,t}$ and $J_{\mathrm{3D}}^{\,t}$, and 2D projection losses ($L_{2\text{D}}$), along with a temporal smoothness regularizer ($L_{\text{Smooth}}$) that discourages frame-to-frame jitter:
\[
\arg\min_{\alpha,\gamma,\phi,\theta}\;
w_{3\text{D}}L_{3\text{D}} + w_{2\text{D}}L_{2\text{D}} + L_{\text{Smooth}}.
\]
We optimize this objective with a Levenberg–Marquardt solver implemented in JAX~\cite{yi2024egoallo}. Running on an NVIDIA A100 GPU, the optimizer processes a 300‑frame sequence in approximately 20 ms after compilation.
Please refer to Sec.~\ref{sec:appendix-real-to-sim} for more details.

\subsection{Generating Simulation-Ready Data}
\label{subsec:meshification}

To deploy the monocular reconstruction in a physics engine, we (i) align it with real-world gravity using GeoCalib~\cite{veicht2024geocalib} and (ii) convert the noisy, dense point cloud into a lightweight mesh that imposes meaningful geometric constraints and supports memory-efficient parallel training. We use NKSR~\cite{huang2023nksr} for meshification. Please see Sec.~\ref{sec:appendix-generating-sim-data} for more details.

\section{Policy Learning}
\vspace{-5pt}
\label{sec:rl}

\begin{figure*}[t]
    \centering
    \includegraphics[width=1.01\linewidth,clip,trim=0cm 0cm 0cm 0cm]{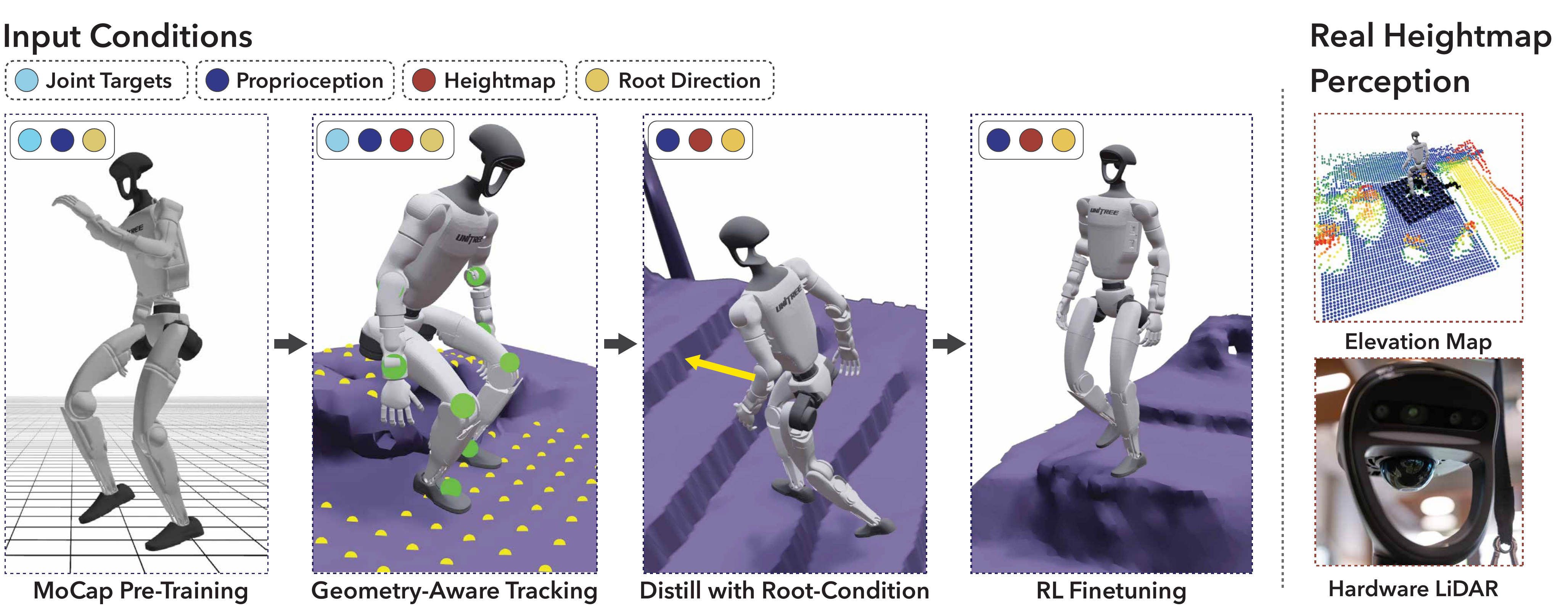}
    \caption[]{\textbf{Policy training in sim.} Our pipeline of training RL starts with a dataset of Motion Capture trajectories. We then inject a heightmap observation and track whole-body reference trajectories from our videos in various environments. We proceed to distill a policy conditioned only on the root position of the robot. We then finetune this policy directly with RL using the same reduced observation set. Our pipeline is motivated by three goals: (a) producing motions that are fast and faithful to the original video demonstrations; (b) ensuring observations are available in real-world settings; and (c) training a generalist policy that distills knowledge from all video demonstrations into a single model applicable beyond the training set. \vspace{-1em}
    }
    \label{fig:rl_method}
\end{figure*}

Given the kinematic reference from our clips and scenes, our policy learning pipeline produces a context-conditioned policy that can perform skills from the references when prompted by the appropriate environmental context. Figure~\ref{fig:rl_method} gives an overview of our pipeline, detailed below. %

\textbf{Policy Learning.} We use Proximal Policy Optimization \cite{ppo} implementation from \citet{rudin22} for training our policy. %
Our learning takes place in the IsaacGym simulator \cite{isaacgym}. 
Please refer to Sec.~\ref{sec:supplimentary:policy-learning} for more details on our formulation and hyperparameters.

\noindent \textbf{Observations.}
Our policies are conditioned on both proprioceptive and target-related observations. The proprioceptive inputs include a history of the robot’s joint positions ($q$), joint velocities ($\dot q$), angular velocity ($\omega$), projected gravity vector ($g$), and previous actions ($a^{t-1}$); we use a history length of 5 in practice.
In addition, the policy receives local target observations: the target joint angles, target root roll and pitch, and the desired root direction, specified by relative x-y offset and yaw angle between the robot’s current root position and the target root, all expressed in the robot’s local frame.
For policies conditioned on heightmaps, we further provide an elevation map around the torso. This is represented as an $11 \times 11$ grid sampled at $0.1\mathrm{m}$ intervals, which captures local terrain geometry.
Finally, the critic receives additional privileged observations, which are detailed in Table~\ref{table:policy-inputs}. 

\noindent \textbf{Batched Tracking.} Our system utilizes a batched variant of DeepMimic \citep{DeepMimic} in order to learn to imitate motions using RL. We implement Reference State Initialization \cite{DeepMimic} in addition to motion load balancing similar to \citet{tessler2024maskedmimic}, upweighting motions with a lower success rate. 

\noindent \textbf{Rewards.} 
Our RL reward is designed entirely around data-driven tracking terms—specifically, link and joint positions, joint velocities, and foot-contact signals---so that raw demonstrations can be translated into physically executable motions with minimal hand-tuning.
We have two objectives: (1) reducing reliance on manually crafted priors that are typically introduced through reward engineering,
and (2) ensuring physical feasibility of the resulting motions.
These two goals can conflict: because the reference trajectories are purely kinematic data from humans, exact tracking may result in non-physical motion. 
We therefore introduce an action-rate penalty along with several other penalty criteria designed to discourage exploiting simulator physics.
Full formulations and weights are detailed in Sec.~\ref{sec:supplimentary:policy-learning:rl}. %
We train our policy over the stages described below.

\noindent \textbf{Stage 1: MoCap Pre-Training.} MoCap pre-training lets a policy learn challenging skills from noisy video reconstruction while keeping hand-crafted priors to a minimum and bridging the human-to-robot embodiment gap.
Earlier work tackles this either by sampling start poses with multi-agent RL \cite{SfV} or by having a privileged simulator imitate the motion \cite{ASAP}.
Radosavovic et al.~\cite{radosavovic2024humanoid} and Singh et al.~\cite{himanshu2024} instead employed a form of kinematic pre‑training on human data.
We adopt a simpler yet effective strategy: first pre‑train the policy on MoCap trajectories, then fine‑tune it on our reconstructed video data---both stages use reinforcement learning in a physics simulator. Even the MoCap‑only policy can be deployed directly on the real robot as shown in Fig~\ref{fig:progressive-deployment}. We used LAFAN motion capture data~\cite{harvey2020robust} retargeted to Unitree G1. For the pretrained policies, the conditioning the policy receives is the target joint angles, target root roll/pitch, and desired root direction.

\noindent \textbf{Stage 2: Scene-Conditioned Tracking.} 
After MoCap pretraining, we initialize the policy from MPT checkpoints and introduce scene awareness by conditioning on the environment heightmap. The heightmap is integrated via a projection into the MPT policy’s latent space residually with an initial weight of 0. %
We then randomly sample motions and perform DeepMimic-style tracking across reconstructed terrains. During this stage, the policy continues to receive motion-specific tracking conditioning, including target joint angles, root roll/pitch, and desired root directions. %

\noindent \textbf{Stage 3: Distillation.} Following the stage of batched tracking, we distill via DAgger \cite{dagger} to a policy that does not observe target joint angles or root roll/pitch observations.  We are then able to use the desired root directions observations as conditioning signals to control the robot's position, which can be fed either from a joystick or potentially a path provided by a high-level controller. %
In this way, our framework unifies the previously separate approaches of joystick tracking and global reference following. Our distilled policy benefits from the fact that the teacher is also trained with observation randomization, hence it learns actions under some uncertainty, which would not be the case if we started by training with full body observations; this has been shown to be helpful in other contexts with policy learning \cite{dextrah}.

\noindent \textbf{Stage 4: Under-conditioned RL Finetuning.} After distilling our policies to be exclusively conditioned on the root of our trajectories, we perform another round of RL. This is because behaviours which are learned conditioned on target joints may be sub-optimal for policies which are not conditioned on such targets. In practice, we found that this can significantly boost performance as compared to distilled policies. It also makes it possible to add lower-quality reference motions to the reference set since removing targets from the actor in effect makes it a ``data-driven" reward signal with an under-constrained actor which is able to follow references appropriate to context.

\begin{figure*}[t!]
  \centering
    \includegraphics[width=1.0\linewidth,clip,trim=0cm 0cm 0cm 0cm]{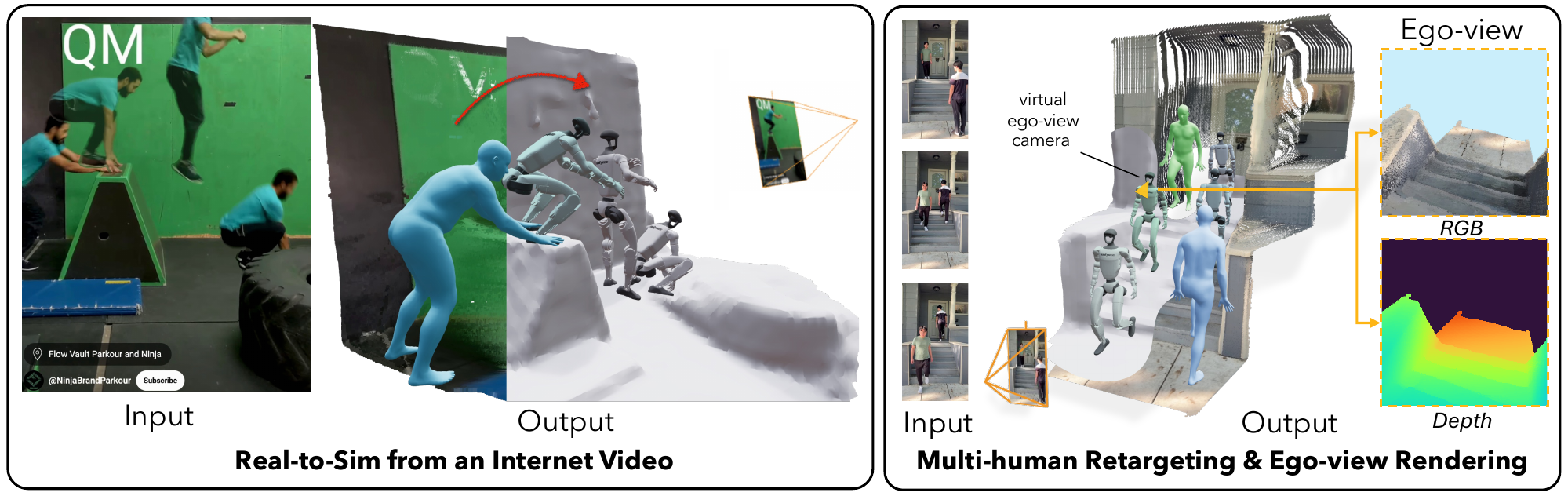}
    \caption[]{\textbf{Versatile capabilities of our Real-to-Sim pipeline.}  
    VideoMimic enables (i) robust tracking of Internet videos with challenging motion and diverse environments,
(ii) simultaneous reconstruction and retargeting of multiple humans, and
(iii) ego-view RGB-D rendering for embodied perception---though not used in our current policy, it highlights the framework’s broader applicability across inputs and tasks.
}
  \vspace{-4mm}
  \label{fig:real2sim_qualitative}  
\end{figure*}

\section{Results}
\vspace{-5pt}
\label{sec:result}
We demonstrate that humanoid robots can learn context-aware skills in diverse environments by imitating everyday human videos. We first evaluate the robustness of our reconstruction pipeline against baselines. Next, we demonstrate its versatility, highlighting its potential impact on future research. We then detail our curated video dataset. Finally, we ablate the MPT component and present demonstrations successfully transferred from simulation policies to a physical robot.

% begin by detailing our curated video data as well as the reconstruction pipeline's versatility and robustness against baselines. We ablate MPT and present demonstrations successfully transferred from simulation policies to a physical robot.

\subsection{Reconstruction and Data}

\noindent\textbf{Evaluation.} We evaluate the robustness of our reconstruction pipeline on a subset of the SLOPER4D dataset~\cite{dai2023sloper4d}, assessing both human trajectory and scene geometry reconstruction.

\begin{wraptable}[6]{r}{0.45\textwidth} %
  \centering
  \resizebox{\linewidth}{!}{%
    \begin{tabular}{lccc}
      \toprule
      Methods & WA-MPJPE & W-MPJPE & Chamfer Distance \\
      \midrule
      WHAM*~\cite{shin2024wham} & 189.29 & 1148.49 & - \\
      TRAM~\cite{wang2024tram} & 149.48 & 954.90 & 10.66 \\
      \midrule
      \textbf{Ours} & \textbf{112.13} & \textbf{696.62} & \textbf{0.75}\\
      \bottomrule
    \end{tabular}
  }
  \caption{\textbf{Comparison of Reconstruction.} *: WHAM does not recover the environment.}
  \label{tab:recon_evaluation}
\end{wraptable}

We compare our method against baselines following the standard evaluation protocol~\cite{shin2024wham, wang2024tram}. 
As summarized in Table~\ref{tab:recon_evaluation}, our method consistently achieves the best performance, outperforming prior work in both human trajectory accuracy (WA/W-MPJPE) and scene geometry (Chamfer Distance). Refer to Sec.~\ref{sec:eval} for more evaluation details.

\begin{figure*}[t]
  \centering

    \makebox[\textwidth][c]{\includegraphics[width=1.075\linewidth,clip,trim=0cm 0cm 0cm 0cm]{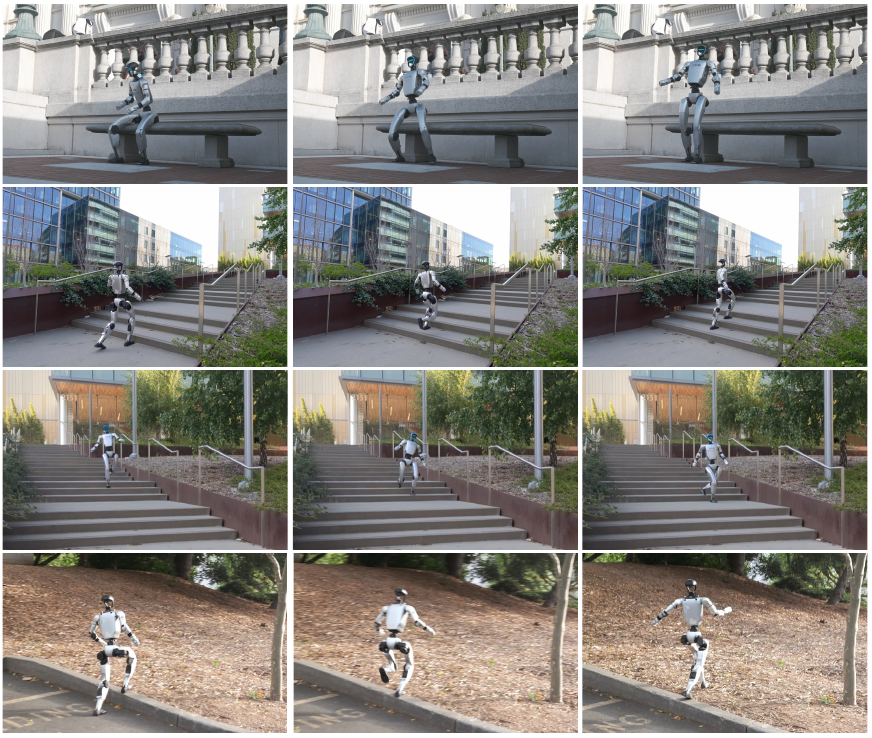}}

  \caption[]{\textbf{The policy performing various skills on the real robot:} traversing complex terrain, standing, and sitting. All these skills are in a single policy, which decides what to do based on the context of its heightmap and joystick direction input. \textit{Top row}: the policy stands from a seated position after sitting down. \textit{Second row}: the policy walks up a flight of stairs. \textit{Third row}: the policy walks down a flight of stairs. \textit{Bottom row}: the policy walks over a kerb and onto a rough terrain. Please find the video results on our \href{https://www.videomimic.net/}{webpage}.}
  \label{fig:real_eval}
  \vspace{-1em}
\end{figure*}

\noindent\textbf{Versatility.}
Figure~\ref{fig:real2sim_qualitative} and our \href{https://www.videomimic.net/page1.html}{webpage}'s Viser~\cite{yi2025viser} visualizations highlight the breadth of our reconstruction pipeline, showcasing (i) robust environment reconstruction from an Internet video involving dynamic human-scene interaction, %
(ii) multi-human reconstruction and retargeting. Furthermore, the dense point cloud reconstruction enables ego-view RGB-D rendering via simple rasterization. While not used in our current policy, this offers a promising direction for future work—especially given the challenges of rendering naturalistic images in simulation.

\noindent\textbf{Video Data.}
We curated 123 casually recorded smartphone videos of people performing everyday activities in diverse indoor and outdoor settings, including sitting, standing up from furniture, walking up/down stairs (even backwards), and stepping onto blocks. See Sec.~\ref{sec:data_distribbution} for more details.

\begin{wrapfigure}{r}{0.5\textwidth}  %
  \centering
  \vspace{-3mm}
  \includegraphics[width=0.92\linewidth,clip,trim=0 0 0 0]{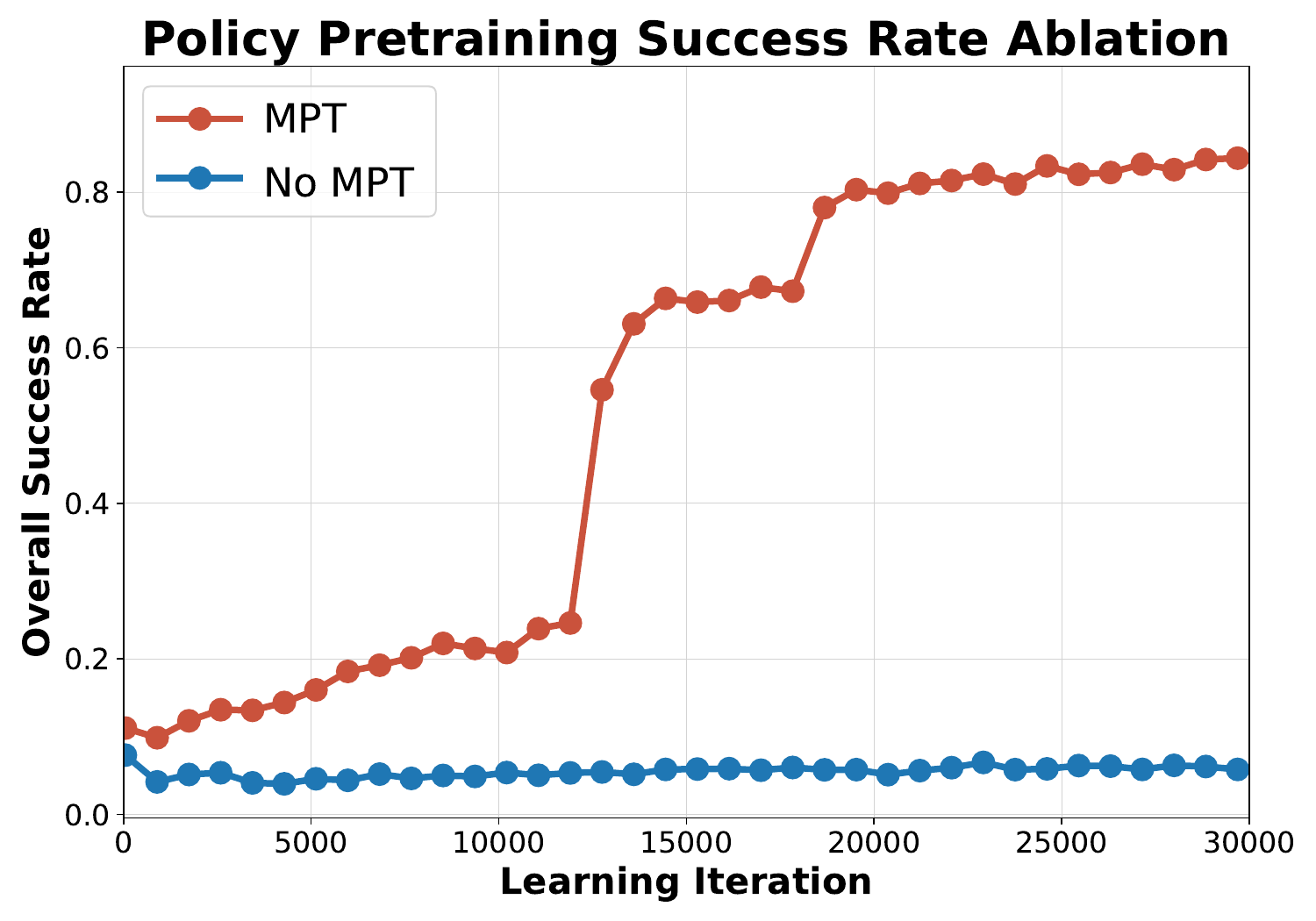}
  \caption{\textbf{Impact of MoCap pre-training (MPT).} Pre-training the policy on motion-capture data facilitates learning on video captures despite noisy references.
 }
 \vspace{-13mm}
  \label{fig:tracking_eval}
\end{wrapfigure}

\noindent \textbf{MPT ablation.} \label{sec:mpt} We ablate the impact of pretraining on motion capture data.
MPT has multiple effects: first, reference motions are noisy and thus harder to learn to track tabula rasa. Second, initial positions of the robot are often not entirely statically stable or may have some interpenetrations with the scene. Hence, MPT can help stabilize learning during the initial phases, whereas a policy from scratch may not even be able to learn how to balance. As shown in Figure~\ref{fig:tracking_eval}, removing MPT significantly hinders the policy’s ability to learn effective behaviors.

\subsection{Real-world Deployment}

\noindent\textbf{Setup.}
We deploy our controller on a 23-DoF \textit{Unitree G1} humanoid and run it onboard at 50 Hz. Following~\cite{zakka2025mujocoplayground}, we set relatively low joint gains, \(K_p = 75\), to avoid excessively fast or overly stiff behaviour—which helps to avoid excessively violent contact when the robot makes heavy contact with objects such as chairs or stairs.
LiDAR odometry and height-maps are computed in real time using Fast-lio2~\cite{xu2021fastlio2fastdirectlidarinertial} and probabilistic terrain mapping~\cite{Fankhauser2018ProbabilisticTerrainMapping, long2024learning}. 
The odometry is used to align the local height-map observations into the robot's frame.
We feed joystick targets from a human operator.
Including policy running, all operations are run onboard.
We found two critical ingredients for successful motion deployment through iterative sim-to-real trials: (i) relaxing the episode-termination tolerances with respect to the reference motion, and  
(ii) injecting realistic physics perturbations during training. Full details are given in Sec.~\ref{sec:appendix:robot-deployment}.

\noindent\textbf{Real-world evaluation.}  
Figure~\ref{fig:real_eval} and the accompanying video showcase the policy executing a wide range of whole-body behaviors on the Unitree~G1.  
Without any task-specific tuning, the same network---driven only by proprioception and a noisy LiDAR height-map that provides a full \(360^{\circ}\) view around the torso---climbs and descends indoor and outdoor staircases, traverses steep earthen slopes and rough vegetation, and reliably sits down on or stands up from chairs and benches.  
The controller is surprisingly resilient: after unexpected foot slides while descending stairs, it recovers by momentarily hopping on a single leg before regaining nominal gait.  

\noindent To the best of our knowledge, this is the \emph{first} real-world deployment of a \emph{context-aware} humanoid policy learned from monocular human videos, jointly demonstrating perceptive locomotion and environment-prompted whole-body skills such as sitting, standing, and climbing stairs.  
Additional qualitative results are available on the \href{https://www.videomimic.net/}{project webpage}.
\section{Conclusion}
\label{sec:conclusion}

We introduced \textsc{VideoMimic}, a real-to-sim-to-real pipeline that converts everyday human videos into environment-conditioned control policies for humanoids. The system (i) reconstructs humans and surrounding geometry from monocular clips, (ii) retargets the motion to a kinematically feasible humanoid, and (iii) uses the recovered scene as task terrain for dynamics-aware RL. The result is a \emph{single} policy that delivers robust, repeatable {contextual control}—e.g., stair ascents/descents and chair sit-stand—all driven only by the environment geometry and a root direction command. 
\textsc{VideoMimic} offers a scalable path for teaching humanoids contextual skills directly from videos. We expect future work to extend the system to richer human–environment interactions, multi-modal sensor-based context learning, and multi-agent behavior modeling, among other directions.

\section{Limitations}
\label{sec:limitations}

Our pipeline delivers encouraging real‑world results, yet several practical weaknesses remain.

\vspace{-0.9em}
\paragraph{Reconstruction.}  
Monocular 4D human–scene recovery is still brittle in the wild. Camera pose drift in MegaSaM often yields duplicate ``ghost" layers of the same surface. Due to its inability to refine the dynamic points, the dynamic points from the person are mistakenly fused into the static point cloud or inaccurately placed (e.g., feet buried beneath the environment). In particular, we found that MegaSaM performs poorly on images with low texture. Depth filtering and spatio‑temporal subsampling remove many outlier points, but aggressive thresholds leave holes that hinder meshing. NKSR mitigates noise, yet may oversmooth fine geometry (e.g., narrow stair treads); such high-frequency details are crucial for robot control, and we discard videos where these details are missing after reconstruction.  
Also, during point-to-mesh conversion, spiky artifacts may appear due to stray points.
\vspace{-2em}
\paragraph{Retargeting.}  
The kinematic optimizer assumes every reference pose can be made feasible once scaled to the robot. In cluttered scenes, this is not always true, and conflicting costs—strict foot‑contact matching versus collision avoidance—can trap the solver in poor local minima that the RL controller must subsequently “clean up.’’
\vspace{-1em}
\paragraph{Sensing and policy input.}  
At test time, the controller receives only proprioception and an \(11\times11\) LiDAR height‑map.  This coarse grid is adequate for terrain and chairs but lacks the resolution for precise contacts, manipulation, or reasoning about overhanging obstacles. Incorporating richer perceptual inputs—such as RGB‑D data or learned occupancy grids—would likely broaden the method’s applicability and improve its semantic understanding of the environment.
\vspace{-1em}
\paragraph{Simulation fidelity.}  
We assume the scene can be represented as a single rigid mesh. Scaling to articulated or deformable objects will require more expressive simulators and object‑level reconstruction pipelines—open problems for future work.
\vspace{-1em}
\paragraph{Data scale and motion quality.}  
The distilled policy is trained on only 123 video clips and occasionally relies on recovery behaviors, leading to jerky motions. Larger, more diverse video corpora and iterative real‑world fine‑tuning should improve smoothness and robustness.

Moving beyond these limitations—through better dynamic static separation, hole‑resistant meshing, adaptive retargeting costs, richer perception, and larger datasets—is a key direction for future work.

\acknowledgments{We thank Brent Yi for his guidance with the excellent 3D visualization tool we use, Viser. We are grateful to Ritvik Singh, Jason Liu, Ankur Handa, Ilija Radosavovic, Himanshu Gaurav-Singh, Haven Feng, and Kevin Zakka for helpful advice and discussions during the paper. We thank Lea Müller for helpful discussions at the start of the project. We thank Zhizheng Liu for helpful suggestions on evaluating human and scene reconstruction. We thank Moji Shi and Huayi Wang for advice on using LiDAR heightmap input. We thank Eric Xu, Matthew Liu, Hayeon Jeong, Hyunjoo Lee, Jihoon Choi, Tyler Bonnen, and Yao Tang for their help in capturing and featuring in the video clips used in this project.
We acknowledge support from the BAIR Humanoid Intelligence Center.
Chung Min Kim is supported by the NSF Research Fellowship Program, Grant DGE 2146752. Pieter Abbeel holds concurrent appointments as a Professor at UC Berkeley and as an Amazon Scholar. This paper describes work performed at UC Berkeley and is not associated with Amazon. This project was funded in part by NSF: CNS-2235013, IARPA DOI/IBC No. 140D0423C0035, ONR MURI awards N00014-21-1-2801 and W911NF-23-1-0277, and Bakar fellows.}

\clearpage
\bibliography{bib}  %

\clearpage

% \clearpage
\appendix

\section{Real-to-Sim Details}
\label{sec:appendix-real-to-sim}

Our goal is to endow a humanoid with whole-body skills—walking, climbing, sitting---that account for the surrounding geometry after watching a collection of monocular RGB videos of humans performing such skills. We assume (i) a monocular RGB video that clearly captures both the person and the scene; (ii) a static environment during the clip so human motion and terrain can be treated as rigid at training time; and (iii) known robot kinematics and joint limits, but no multi-view rig or motion-capture setup, depth sensors, or pre-scanned meshes. From the video, we jointly reconstruct metric-scale 4D human trajectories and dense scene geometry, align them to gravity, meshify scene point clouds, and retarget the kinematics to the robot while enforcing contacts and collisions. The resulting motion-and-mesh pairs serve as simulator-ready training clips.

\subsection{Notations}
\label{subsec:preliminary_notation}

\noindent \textbf{Setup.} 
Our method takes a monocular video sequence as input. We denote each video frame as $I^{t}$, with resolution $H \times W$. 
Given this input, along with initial estimates of camera parameters and human poses, our method jointly reconstructs global human motion trajectories and dense environmental geometry in a metric-scale 3D world.

\noindent \textbf{Human.} We represent reconstructed humans from a video using the SMPL model \cite{loper2023smpl}. SMPL is a differentiable function mapping pose parameters $\theta \in \mathrm{SO}(3)^J$ and shape parameters $\beta \in \mathbb{R}^{B}$ to a mesh with $J$ joints. The mesh is positioned in the world coordinate system by global orientation $\phi \in \mathrm{SO}(3)$ and translation $\gamma \in \mathbb{R}^{3}$. Thus, at frame $t$, a human is defined by:
\begin{equation}
\textbf{P}^t = {\phi^t, \theta^t, \beta^t, \gamma^t}.
\end{equation}

\noindent \textbf{Camera.} We assume a perspective camera model with intrinsics \(K \in \mathbb{R}^{3 \times 3}\) and extrinsics defined by rotation \(R \in \mathrm{SO}(3)\) and translation \(t \in \mathbb{R}^{3}\). A 3D point \(x^{\text{3D}} \in \mathbb{R}^3\) is first transformed into the camera frame and then projected onto the image plane as:
\begin{equation}
x_{\text{2D}} = \Pi\left(K \begin{bmatrix} R & t \end{bmatrix} \begin{bmatrix} x_{\text{3D}} \\ 1 \end{bmatrix} \right),
\end{equation}
where \(\Pi: \mathbb{R}^3 \to \mathbb{R}^2\) denotes the perspective projection, defined as \(\Pi\left( \begin{bmatrix} u \\ v \\ w \end{bmatrix} \right) = \left( \frac{u}{w}, \frac{v}{w} \right)\).

\noindent \textbf{Scene.} The scene is represented using dense per-frame depth and camera output by MegaSaM \cite{li2024megasam} or MonST3R~\cite{zhang2024monst3r}. Specifically, we use MegaSaM for scenes with richer textures where the correspondence-based Bundle-Adjustment is more reliable. For textureless scenes, we adopt MonST3R and its depth-conditioned variant~\cite{lu2024align3r} for better reconstruction results, although MegaSam generalizes better across a wider distribution of videos.
To resolve scale ambiguity of the camera translation and scene geometry, we introduce a scaling parameter $\alpha$. Given per-pixel depth $D_{i,j}$, the corresponding world coordinate $\mathcal{S}{i,j}$ for pixel $(i,j)$ is obtained by unprojecting the points:
\begin{equation}
\mathcal{S}{i,j} = \alpha (R^\top [K^{-1}D_{i,j}[i,j,1]^\top] - R^\top t).
\end{equation}

\begin{figure*}[t]
  \centering
  % --- wrap the two sub‑figures in a TikZ picture ----------------------------
  \begin{tikzpicture}[remember picture]
    % put both sub‑figures inside one minipage so TikZ knows their bounds
    \node (block) [inner sep=0] {%
      \begin{minipage}{\textwidth}
        \begin{subfigure}{0.49\linewidth}
          \includegraphics[width=\linewidth,clip]{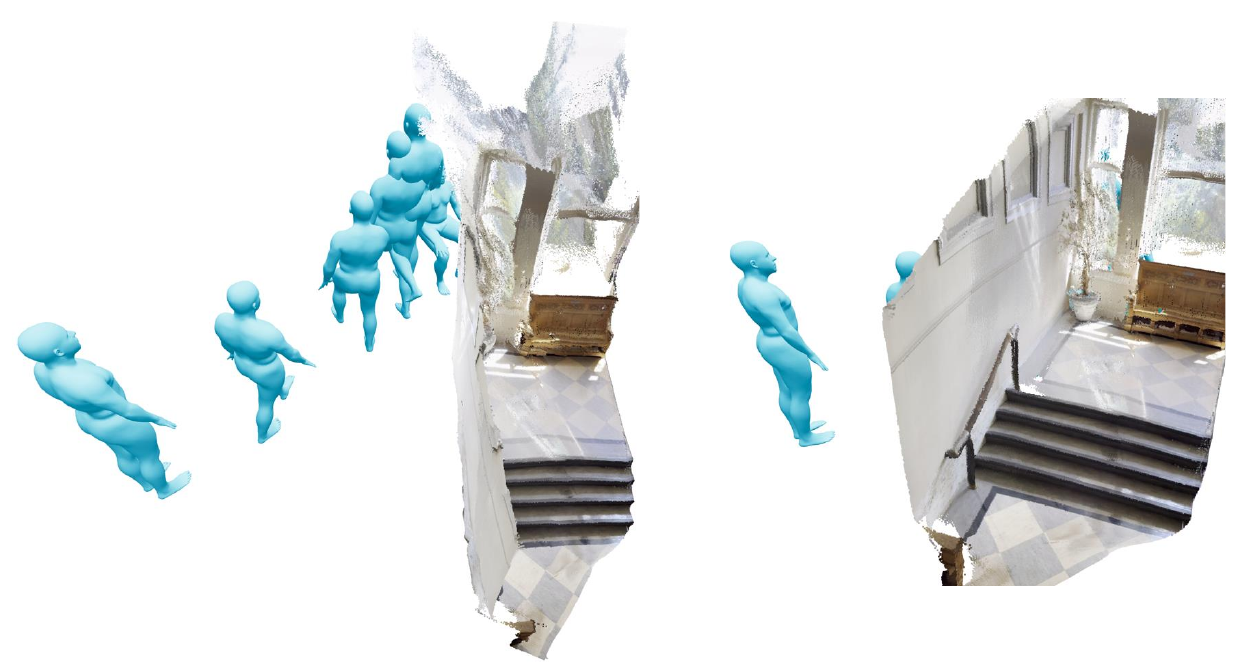}
          \caption{Before}
          \label{fig:before_optimization}
        \end{subfigure}\hfill
        \begin{subfigure}{0.49\linewidth}
          \includegraphics[width=\linewidth,clip]{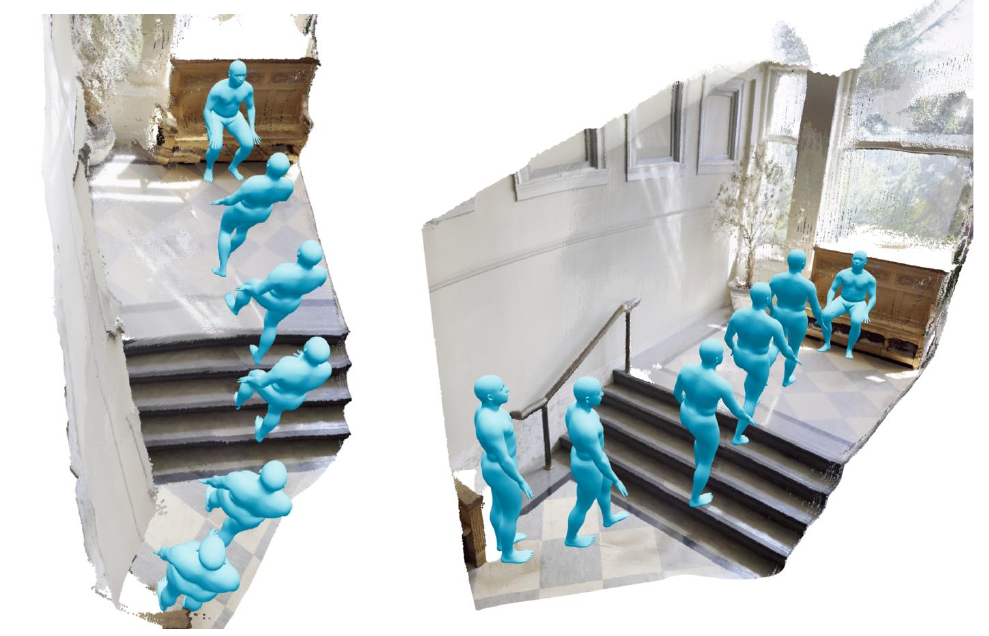}
          \caption{After}
          \label{fig:after_optimization}
        \end{subfigure}
      \end{minipage}%
    };

    % --- draw the dotted line -------------------------------------------------
    % X‑coordinate: midway between the two sub‑figures  (= centre of block)
    \path (block.north) -- (block.south)
          coordinate[midway] (mid);     % ‘mid’ is halfway down
    \draw[dotted, line width=0.4pt] (mid |- block.north) -- (mid |- block.south);
  \end{tikzpicture}
  % ---------------------------------------------------------------------------

  \caption{\textbf{Before and after optimization.} We visualize human trajectories and scene point clouds before and after optimization, showing each from two different viewpoints.}
\end{figure*}

\subsection{Objectives for Joint Human–Scene Reconstruction}
As shown in Figure~\ref{fig:before_optimization}, the coarse initialization produces inaccurate human trajectories and an incorrect scene scale. Our optimization procedure refines these estimates by jointly optimizing the human’s global translations ($\gamma^{1:T}$), global orientations ($\phi^{1:T}$), local poses ($\theta^{1:T}$), and the scene point cloud scale ($\alpha$). The optimization aligns the human and world as shown in Figure~\ref{fig:after_optimization}, and aids in generating simulation-ready data.

The objective combines joint‑distance losses in 3D and through 2D projection with a temporal‑smoothness regularizer, which discourages frame‑to‑frame jitter in both the root trajectory and the articulated pose:
\[
\arg\min_{\alpha,\gamma,\phi,\theta}\;
w_{3\text{D}}L_{3\text{D}} + w_{2\text{D}}L_{2\text{D}} + L_{\text{Smooth}}.
\]

\begin{align}
L_{3\text{D}} &= \sum_{t}\bigl\lVert J_{\mathrm{3D}}^{\,t}-\tilde{J}_{\mathrm{3D}}^{\,t}\bigr\rVert_{1},\\[2pt]
L_{2\text{D}} &= \sum_{t}\bigl\lVert
  \tilde{J}_{\mathrm{2D}}^{\,t}
  -\Pi\!\Bigl(
      K\!\begin{bmatrix} R^{t} & t^{t} \end{bmatrix}
      \begin{bmatrix} J_{\mathrm{3D}}^{\,t} \\ \mathbf{1} \end{bmatrix}
    \Bigr)
  \bigr\rVert_{1},\\[2pt]
L_{\text{Smooth}} &= \lambda_{\gamma}\sum_{t}\bigl\lVert\gamma^{t}-\gamma^{t-1}\bigr\rVert_{2}^{2}
                + \lambda_{\theta}\sum_{t}\bigl\lVert\theta^{t}-\theta^{t-1}\bigr\rVert_{2}^{2},
\end{align}

where \(\Pi: \mathbb{R}^3 \to \mathbb{R}^2\) denotes the perspective projection, defined as \(\Pi\left( \begin{bmatrix} u \\ v \\ w \end{bmatrix} \right) = \left( \frac{u}{w}, \frac{v}{w} \right)\). 
We optimize this objective with a Levenberg–Marquardt solver implemented in JAX~\cite{yi2024egoallo}. Running on an NVIDIA A100 GPU, the optimizer processes a 300‑frame sequence in approximately 20 ms after compilation.

\subsection{Generating Simulation Ready Data} \label{sec:appendix-generating-sim-data}

\textbf{Gravity alignment.}
We first estimate the gravity direction in the initial camera frame and define the transformation that converts the 3D reconstruction from world coordinates to a gravity-aligned coordinate system compatible with physics engines. GeoCalib \cite{veicht2024geocalib} provides the roll–pitch angles for that frame; the composite rotation \(R_{\text{gravity,world}}\!=\!R_{y\uparrow\!\to z}R_{x}(\pi)R_{z}(\rho)R_{\text{cam,world}}\) re-orients the reconstruction so that \(+z\) points up, ensuring consistency with gravity in physics engines. The transformation is applied to both human keypoints and static scene geometry.

\textbf{Pointcloud filtering.}  
For each world pointcloud we discard noisy background points and human pointclouds with thresholding on depth gradient, rotate by \(R_{\text{gravity,world}}\), scale, and crop to a \(\pm2\mathrm{\,m}\) box around the SMPL joints per frame. A \(0.1\mathrm{\,m}\) voxel grid then keeps at most 20 samples per occupied cell, shrinking the pointcloud to about 5\% of its original size without losing surface detail.

\textbf{Meshification.}  
We first surface it using NKSR~\cite{huang2023nksr} to obtain a coarse mesh. Top-down ray casting fills large holes via inverse-distance interpolation inside the convex hull; merging the infilled points with the originals and re-running NKSR produces the final mesh. The entire pipeline processes a 300-frame sequence in roughly 60\,s  
(\(\sim\!5\) s for gravity alignment and \(\sim\!55\) s for filtering and meshing).

\textbf{Humanoid Motion Retargeting.}
\label{subsec:retargeting}
Given the human trajectories and environment meshes as input, our objective is to transform these motions to the \textit{robot}'s embodiment. 
We approach the retargeting task as an optimization problem similar to previous work~\cite{h2o, omnih2o}, but instead use a Levenberg-Marquardt solver to handle the highly nonlinear landscape of the human-in-scene problem setting. 

We solve for the following variables: the G1 joint angles $q^{1:T}$, its root poses $(\phi_\text{R}^{1:T}, \gamma_\text{R}^{1:T})$, and the set of per-link scale factors $s$ between the two embodiments. Note that $s$ is \textit{constant} across timesteps, and for all distance-related costs we penalize $x$, $y$, and $z$ components independently (i.e., no norm).

Inspired by~\citet{cheynel2023sparse}, we implement a kinematic tree-based motion transfer cost $L_{\text{motion}}^{t} = L_{\text{position}}^{t} + L_{\text{angle}}^{t}$:

\begin{equation}
L_\text{position}^{t} = \sum_{i \ne j} m_{ij} \left\| \Delta_{ij}^\text{SMPL}(t) - s \cdot \Delta_{ij}^\text{G1}(t) \right\|_2^2,
\end{equation}

\begin{equation}
L_\text{angle}^{t} = \sum_{i \ne j} m_{ij} \left( 1 - \left\langle \Delta_{ij}^\text{SMPL}(t),\ \Delta_{ij}^\text{G1}(t) \right\rangle \right),
\end{equation}

where $\Delta_{ij}$ is the position difference between joints $i$ and $j$. $m_{ij}$ denotes if the joints are immediate neighbors in the robot's kinematic chain --- 1 if true, 0 if not. These two costs are weighed equally. These terms are regularized to be close to $1.0$ and stay non-negative. 

We also ground the robot with the environment through a set of contact costs (e.g., foot contact point matching, foot skating penalty) and collision costs (self- and world-collision avoidance). The robot is approximated as a set of capsules, and the world as a heightmap. The learning-based feet contact estimation~\cite{huang2022rich} is used to get contact signals. 

The skating cost is defined as:

\begin{equation}
L_{\text{skating}}^t = \sum_{\text{foot} \in \{\text{L}, \text{R}\}} 
% \Delta f_\text{foot}^t 
\left\| p_{\text{foot}}^{t} - p_{\text{foot}}^{t-1} \right\|
+ \left\| p_{\text{ankle}}^{t} - p_{\text{ankle}}^{t-1} \right\|,
\end{equation}

where $p_{\text{foot}}^{t} = \mathbf{T}_{\text{root}, \text{world}}(\phi_R^t, \gamma_R^t) 
 \mathbf{T}_{\text{foot}, \text{root}}(q^t)$ is the position of the foot link in the world at time $t$.

We also include common robot costs (e.g., joint limits, temporal smoothness of root pose and robot joints) and a small regularization cost on the G1 robot's knee yaw joint for stable leg poses. %
Processing a 300-frame clip takes around 10 seconds on a single A100, using the PyRoki library~\cite{pyroki2025}. 
% Please see the supplement for more details. 

\subsection{Evaluation Details}
\label{sec:eval}
% For quantitative evaluation of our real-to-sim for human trajectory and geometry reconstruction, we evaluate on SLOPER4D dataset. 
% We follow the convention~\cite{liu2025josh, shin2024wham, wang2024tram} to evaluate the human trajectory with World-Aligned mean per joint position error (WA-MPJPE) and World coordinate (W-MPJPE) mean per joint position error, and evaluate the reconstruction error with the Champer distance. 
For quantitative evaluation of our real-to-sim pipeline, we conduct experiments on the SLOPER4D dataset~\cite{dai2023sloper4d}. 
Following established protocols~\cite{liu2025josh, shin2024wham, wang2024tram}, we assess human trajectory reconstruction using World-frame Mean Per Joint Position Error (W-MPJPE) and World-frame Aligned  MPJPE (WA-MPJPE), and evaluate scene geometry reconstruction using the Chamfer Distance. 
Specifically, for human trajectory evaluation, we first slice each sequence into 100-frame segments. W-MPJPE is then computed by aligning only the first two frames to the ground truth, emphasizing global consistency, while WA-MPJPE aligns the entire segment to evaluate local trajectory accuracy.
For geometry evaluation, we compute the Chamfer Distance (in meters) between the predicted pointcloud and LiDAR points within the RGB camera's field of view. 
For benchmarking, we use a subset of SLOPER‑4D that contains only those sequences in which SAM2 tracking—human detection plus cross‑frame association—succeeds. This yields two sequences each of running, walking, and stair ascent/descent. Results for WHAM and TRAM are reproduced with their official code.

\begin{figure*}[t]
  \centering
    \includegraphics[width=1.0\linewidth,clip,trim=0cm 0cm 0cm 0cm]{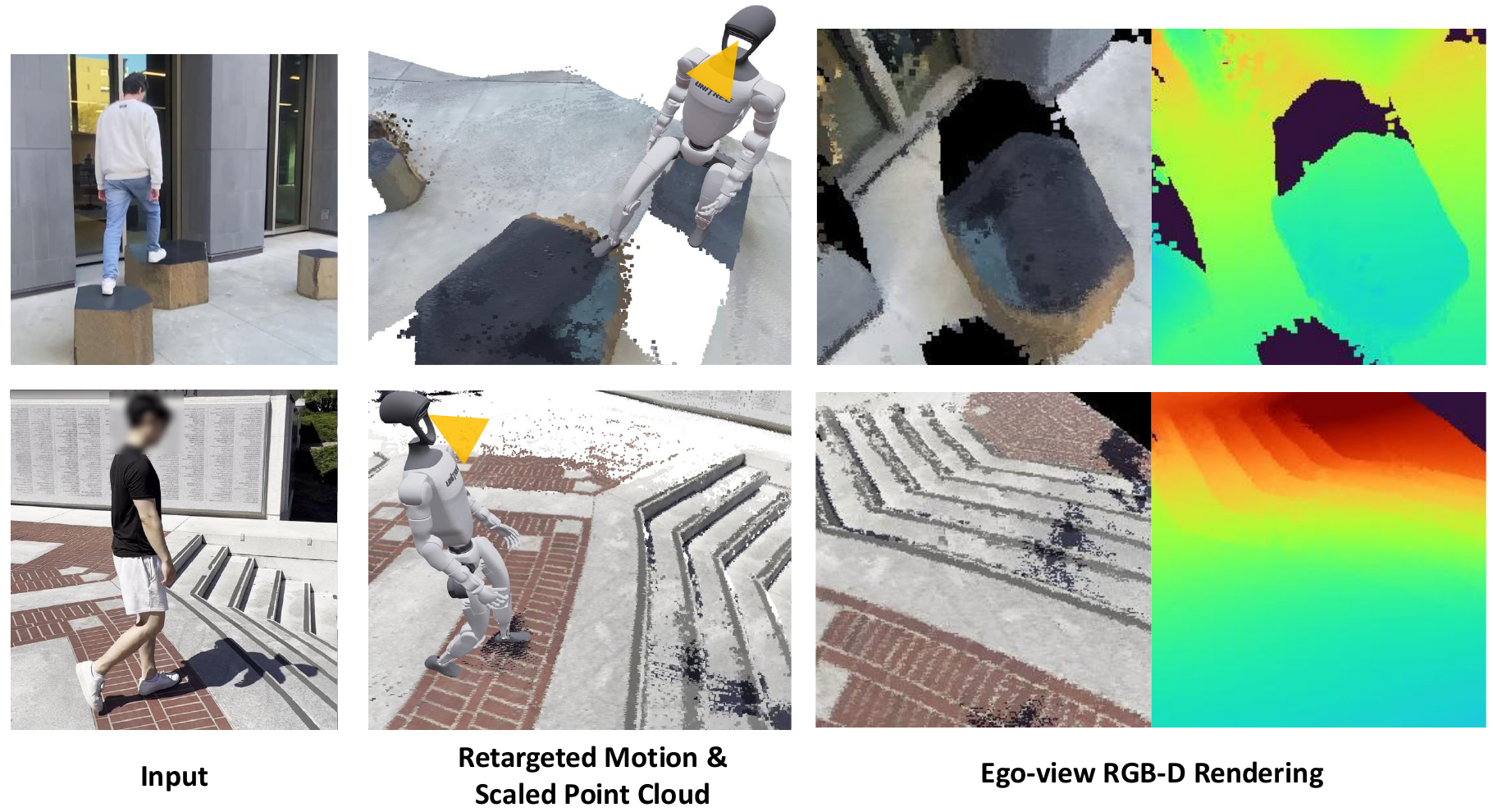}
    \caption{\textbf{Ego-view RGBD-rendering.} While our policy does not use RGB conditioning for now, we demonstrate the potential of our reconstruction for ego-view rendering. We rasterize the metrically‑scaled point cloud into color and depth images by projecting every 3D point onto the image plane of a virtual camera pitched to match the G1’s head‑mounted sensor. This first‑person rendering gives the robot realistic observations of its surroundings, unlocking future work on active vision and semantic understanding, and can be injected directly into the simulator during RL training to couple perception and control.}
    \label{fig:egoview_rendaring}
\end{figure*}

\subsection{Ego-view Rendering}
We leverage our metrically‑scaled 4D reconstruction to render ego‑view RGB‑D frames by projecting every 3D point into a virtual camera at the Unitree G1 head sensor. This first‑person perspective is not consumed by the policy in the present work, yet it demonstrates the pipeline’s future potential. Our reconstructions can supply realistic visual observations to future perception‑conditioned policies, enabling active vision and semantic scene understanding directly from the robot’s viewpoint. A key limitation, however, is that monocular capture leaves many surfaces unobserved; when rasterized, these occluded regions manifest as holes or gaps in the rendered images. Bridging these gaps is an exciting direction for follow‑up work: modern data-driven novel‑view‑synthesis models can hallucinate plausible geometry and appearance for invisible surfaces, promising photorealistic egocentric renderings that would further narrow the sim‑to‑real perceptual gap for visuomotor policies. 

\section{Policy Learning}
\label{sec:supplimentary:policy-learning}

\subsection{RL Setup and Training}
\label{sec:supplimentary:policy-learning:rl}

We use the PPO algorithm \citep{ppo}. We use a discount factor of $\gamma=0.99$, and a GAE parameter of $\lambda=0.95$. We use an adaptive learning rate with a desired KL of $0.02$. For adaptive learning rate, we use $1.2$ as opposed to the usual $1.5$ learning rate change, as we find that this leads to faster training. When doing RL from scratch, we set the learning rate (which is modified from this value by adaptive KL) at the start to $1e-3$ however when finetuning we find it is important to start from a very low learning rate of $2e-5$ to prevent early updates with a high learning rate from destroying the existing checkpoint when the distribution of data is biased early during an episode when all environments start. We use a max gradient norm of 1.0. 5 learning epochs are used per rollout, and our rollout length is 24. Our entropy coefficient is set to $0.0025$. We use 2 x NVIDIA 4090 GPUs for training, each with 4096 environments (so an effective total of 8192 environments). We use MLPs with 4 layers of [1024, 512, 256, 128] dimensions for different layers.

\subsection{Observations}

Table~\ref{table:policy-inputs} details the actor and critic observations. We note that, unlike works in graphics \cite{DeepMimic, SfV}, the actor gets all unprivileged observations that are all available on the real robot (see Sec.~\ref{sec:result}).
Furthermore, the target root reference during train time is fed into the global frame to ensure it is aligned with the terrain. Thus, unlike \cite{ASAP, exbody2}, we are able to train policies using a global root reference, allowing our policies to correctly imitate references in the world frame over long horizons, which is critical for correctly imitating motions tied to a terrain.  

Fast-lio2~\cite{xu2021fastlio2fastdirectlidarinertial} and probabilistic terrain mapping~\cite{Fankhauser2018ProbabilisticTerrainMapping, long2024learning} are used to derive heightmaps and a live odometry signal in real. This can be used to follow 2D commands in the global frame as we did in testing (see Figure \ref{fig:progressive-deployment}) and as is done in a concurrent work \citep{li2025cloneclosedloopwholebodyhumanoid} for the purpose of teleoperation. However, we note that we do not do this in the final version of the paper to avoid any hysteresis in the position commands, and instead multiply the joystick value by a fixed amount to obtain a position which is interpreted by the policy as an offset to the target position in the local frame.

\begin{table}[!ht]
    \centering
    \resizebox{0.98\linewidth}{!}{%
    \begin{tabular}{@{}lccccc@{}}
    \toprule
    \rowcolor[HTML]{E6E6E6}
    \textbf{Input} & \textbf{Dim.} & \textbf{Actor (MPT)} & \textbf{Actor (tracking)} & \textbf{Actor (distill/ft.)} & \textbf{Critic} \\ 
    \midrule
    Base angular velocity (last 5)               & $5\times3=15$D            & \checkmark & \checkmark & \checkmark & \checkmark \\
    Base linear velocity (last 5)                & $5\times3=15$D            & $\times$   & $\times$   & $\times$   & \checkmark \\
    Projected gravity (last 5)                   & $5\times3=15$D            & \checkmark & \checkmark & \checkmark & \checkmark \\
    DoF positions (last 5)                       & $5\times n_{\mathrm{dof}}$& \checkmark & \checkmark & \checkmark & \checkmark \\
    DoF velocities (last 5)                      & $5\times n_{\mathrm{dof}}$& \checkmark & \checkmark & \checkmark & \checkmark \\
    Actions (last 5)                             & $5\times n_{\mathrm{act}}$& \checkmark & \checkmark & \checkmark & \checkmark \\
    Local frame pos. to target (last 5)          & $5\times2=10$D            & \checkmark & \checkmark & \checkmark & \checkmark \\
    Local frame yaw to target (last 5)           & $5\times1=5$D             & \checkmark & \checkmark & \checkmark & \checkmark \\
    Target joint angles                          & $n_{\mathrm{dof}}$        & \checkmark & \checkmark & $\times$    & \checkmark \\
    Target root roll                             & 1D                        & \checkmark & \checkmark & $\times$    & \checkmark \\
    Target root pitch                            & 1D                        & \checkmark & \checkmark & $\times$    & \checkmark \\
    Terrain height (height-map)                  & 121D                      & $\times$   & \checkmark & \checkmark & \checkmark \\ 
    \midrule
    Root height                                  & 1D                        & $\times$   & $\times$   & $\times$   & \checkmark \\
    Link heights                                 & $n_{\mathrm{links}}$      & $\times$   & $\times$   & $\times$   & \checkmark \\
    Root orientation (quat)                      & 4D                        & $\times$   & $\times$   & $\times$   & \checkmark \\
    Root position                                & 3D                        & $\times$   & $\times$   & $\times$   & \checkmark \\
    Joint positions                              & $n_{\mathrm{dof}}$        & $\times$   & $\times$   & $\times$   & \checkmark \\
    Link positions                               & $3\,n_{\mathrm{links}}$   & $\times$   & $\times$   & $\times$   & \checkmark \\
    Link velocities                              & $3\,n_{\mathrm{links}}$   & $\times$   & $\times$   & $\times$   & \checkmark \\
    Feet contact flags                           & $n_{\mathrm{feet}}$       & $\times$   & $\times$   & $\times$   & \checkmark \\
    Feet target contact flags                    & $n_{\mathrm{feet}}$       & $\times$   & $\times$   & $\times$   & \checkmark \\
    \bottomrule
    \end{tabular}%
    }
    \vspace{0.5em}
    \caption{\textbf{Observations for each network}: Motion Capture Pretraining (MPT) actor, motion-tracking actor, distillation/finetuning actor, and critic.  Here $n_{\mathrm{dof}}=23$ for the Unitree G1.}
    \label{table:policy-inputs}
    \vspace{-1em}
\end{table}

\subsection{Actions}

We use unbounded actions without activations on the actor output following \citep{rudin22}, however following \cite{rl-games2021} we clip the actions to our chosen range prevent the policy from learning bang-bang style control earlier in training and add a "bounds loss" to the actor with the following formulation to enforce this limit in the actor output:
\begin{align}
\tilde\mu_{j}
&= \operatorname{clip}\bigl(\mu_{j},\,-\epsilon,\,\epsilon\bigr)
, \\[0.5em]
\mathcal{L}_{\mathrm{bounds}}
&= \frac{1}{D}\sum_{j=1}^{D}
\bigl|\tilde\mu_{j} - \mu_{j}\bigr|,
\end{align}
where the $\mu_j$ are the policy actions per-dimension. We use a coefficient of $0.0005$ for this loss. For the G1 humanoid, we limit the action magnitude to 8.0. During experiments, our initial action standard deviation is set to 0.8. 

\subsection{Simulation Parameters}

We run Isaac Gym \citep{isaacgym} at 200Hz with control decimation 4, giving an effective policy $\Delta t$ of $0.02s$.

We set the maximum depenetration velocity especially low -- to 0.1m/sec -- to prevent large velocities being applied in the occasional case of interpenetration with terrain during the reference motion.

\subsubsection{Dealing with Terrains in the Simulator}

Due to an undocumented bug in IsaacGym, the simulator registers collisions between robots in different virtual and theoretically isolated ``environments" but which occupy the same world-frame spatial volume. This leads to a dramatic increase in memory usage and a significant drop in simulator throughput. To mitigate this, when training on a single clip or a few clips, we duplicate the task terrain and spatially distribute robot spawns to reduce inter-robot collisions.
% This causes memory usage to dramatically increase and simulator throughput to plummet. In order to resolve this, when we are training on only a single clip or a few clips, we duplicate the task terrain multiple times and distribute spawn of the robots to reduce the number of collisions registered between robots.

\subsection{Reward Formulation}

Our concern above all is to have a system that can take in data and produce new physical motions; hence, our reward is made up mostly of data-driven terms --- tracking link and joint positions and velocities, as well as matching feet contact, with minimal re-weighting. In reward design,
We have two objectives: firstly, to minimize the strength of manually-designed priors injected into RL tracking via reward engineering.\footnote{The more such human priors that are added, the less general our motion tracking system will be. This is the inverse of classical sim-to-real RL pipelines, which largely rely on a practitioner's ``reward hacking" to get desired behaviours.} Secondly, to produce physically feasible motions. 
Note that the two objectives are in tension as since the kinematic data comes from humans, perfect tracking would result in nonphysical motions. We also have several penalty criteria, namely action rate penalties, and two penalties to discourage exploiting simulator physics. Table \ref{table:reward} details the reward terms used for policy learning.

\begin{table}[!htbp]
% \vspace{-2pt}
\centering
\resizebox{0.975\linewidth}{!}{%
\begin{tabular}{l|c|c|c|c}
\toprule
\rowcolor[HTML]{E6E6E6}
Reward & Formula & Weight & $k$ & Justification \\
\midrule
\multicolumn{5}{l}{\textbf{Penalties / Regularization}} \\ \hline
Action‐Rate Penalty      & $\lVert \mathbf{a}_{t}-\mathbf{a}_{t-1}\rVert^{2}$ & $-8.0$  & --   & Can’t move too quickly on real robot                \\
Ankle Action Penalty     & $\lVert \mathbf{a}_{t}\rVert^{2}$                  & $-4.0$  & --   & Ankles should stay near neutral                     \\
DOF-Pos Limits Penalty   & $\displaystyle\sum[\text{overflow w.r.t.\;0.98\,range}]$ & $-50.0$ & -- & Penalize joints outside 98 \% of their range        \\
Collision Penalty        & $\displaystyle\sum 1\!\bigl[\lVert\mathbf{f}_{c}\rVert>0.1\bigr]$ & $-1.0$ & --   & Discourage collisions on penalized links            \\
Contact–No–Velocity Pen. & $\lVert \mathbf{v}_{\text{feet}}\rVert_{(c>1)}$    & $-100$ & --   & Penalize stationary foot contact                    \\
\midrule
\multicolumn{5}{l}{\textbf{Tracking Rewards}} \\ \hline
Joint-Pos Tracking       & $\exp\!\bigl(-k\,\lVert\mathbf{q}-\mathbf{q}^{\ast}\rVert^{2}\bigr)$            & $120$ & $2.0$  & Follow reference joint angles        \\
Joint-Vel Tracking       & $\exp\!\bigl(-k\,\lVert\dot{\mathbf{q}}-\dot{\mathbf{q}}^{\ast}\rVert^{2}\bigr)$ & $24$  & $0.01$ & Follow reference joint velocities    \\
Root Ori Tracking        & $\exp\!\bigl(-k\,\theta_{\text{root}}\bigr)$                                       & $15$  & $3.0$  & Keep base orientation on track       \\
Torso Pos Tracking       & $\exp\!\bigl(-k\,\lVert\mathbf{p}_{\text{torso}}-\mathbf{p}^{\ast}\rVert^{2}\bigr)$ & $15$ & $50.0$ & Precise torso translation            \\
Torso Ori Tracking       & $\exp\!\bigl(-k\,\theta_{\text{torso}}\bigr)$                                      & $15$  & $3.0$  & Precise torso attitude               \\
Link Pos Tracking        & $\exp\!\bigl(-k\,\lVert\mathbf{p}_{\text{links}}-\mathbf{p}^{\ast}\rVert^{2}\bigr)$ & $30$ & $5.0$  & Track 13 key link positions          \\
Link Vel Tracking        & $\exp\!\bigl(-k\,\lVert\dot{\mathbf{p}}_{\text{links}}-\dot{\mathbf{p}}^{\ast}\rVert^{2}\bigr)$ & $5$ & $0.1$  & Track link velocities                \\
Feet Contact Match       & $\sum\mathbbm{1}\!\bigl[c_{t}^{(i)}=c_{t}^{\ast(i)}\bigr]$                         & $1$   & --     & Match contacts from reference motion (BSTRO) \\
Feet Air-Time Bonus      & $\displaystyle\sum_{i}(t^{(i)}_{\text{air}}-k)\,\mathbbm{1}_{\text{first\_contact}}^{(i)}$ & $2000$ & $0.25$ & Reward long, ballistic steps          \\
\midrule
\rowcolor[HTML]{E6E6E6}
Reset & Condition & Weight & $k$ &  \\ \hline
Termination Penalty      & episode end (failure) & $-500$ & -- & Harsh penalty on failure events \\
Alive Reward & added every step & $300$ & -- & Incentive to stay alive \\
\bottomrule
\end{tabular}}
\vspace{1em}
\caption{\textbf{Reward terms}, corresponding weights, and scaling factors $k$. These rewards remain the same across training phases. However, we anneal the action rate and ankle action penalty from 0.2 and 0.0, respectively, to the given values in the table.}
\label{table:reward}
\end{table}

\subsection{Termination Criteria}

Our termination criteria depend on the maximum error of the tracked link joints in the robot. The episode terminates if the Cartesian error exceeds the termination threshold during any step in training. During MPT, we set this to 0.3. When doing the tracking stage over terrains, we set this to 0.5 (higher is better on sometimes noisy references from vision). During RL finetuning, we set this to 1.2. The reason for the high threshold in finetuning is that during RL, we care a lot about ensuring both recovery behaviours and diversity which are not seen during normal DeepMimic-style training on a dataset of the size we are using (and indeed, with too strict a tolerance recovery behaviours will not be seen at all). Hence, a loose tolerance on termination while still using the same tracking reward helps to maintain the essence of the motions in the data while still achieving strong recovery behaviours.

\subsection{Domain Randomization}
\label{subsec:domain_randomisation}

We add various domain randomizations in the simulation to simulate the impacts of various unmodeled physical effects in the simulator, which leads to more robust policies. These are detailed in Table \ref{table:domain-randomisation}.

\begin{table}[!ht]
\centering
\resizebox{1\linewidth}{!}{%
\begin{tabular}{@{}lclp{5.5cm}@{}}
\toprule
\rowcolor[HTML]{E6E6E6}
\textbf{Category} & \textbf{Parameter} & \textbf{Value / Range} & \textbf{Comment} \\ \midrule
\multicolumn{4}{l}{\textbf{Dynamics randomization}} \\ \hline
DoF friction        & $\mu_{\text{DoF}}$          & $\mathcal{U} [0,\,0.02]$            & Different per environment. \\ \midrule
Random pushes (xy)  & $\Delta v_{\text{push},xy}$ & $\mathcal{U}[-0.25,\,0.25]$\,m/s    & Additive to the robot’s reference state or current velocity. \\
Random pushes (z)   & $\Delta v_{\text{push},z}$  & $\mathcal{U}[-0.10,\,0.10]$\,m/s    & Additive to the robot’s reference state or current velocity. \\
Random push interval& $T_{\text{push}}$           & $10$\,s                 & Interval between random pushes. \\ \midrule
\multicolumn{4}{l}{\textbf{Observation \emph{offset} noise (additive bias, re-sampled from gaussian per episode with given standard deviation)}} \\ \hline
Gravity bias        & $\Delta g$                  &  $0.01$g     & Fixed bias added to gravity vector for an episode in policy input. \\
DoF-position bias   & $\Delta q$                  & $ 0.005$rad & Fixed bias added to each joint position in policy input. \\ \midrule
\multicolumn{4}{l}{\textbf{Observation \emph{white} noise (re-sampled every step from Gaussian with divergent standard deviation)}} \\ \hline
Relative odom\,($x,y$) to target   & $\sigma_{xy}$ & $0.01$\,m         & Resampled every step. \\
Relative odom\,yaw to target       & $\sigma_{\psi}$ & $0.01$\,rad       &  \\
DoF positions                       & $\sigma_q$      & $0.01$\,rad       &  \\
DoF velocities                      & $\sigma_{\dot q}$ & $1.5$\,rad/s     &  \\
Linear base velocity                & $\sigma_{v}$     & $0.1$\,m/s        &  \\
Angular base velocity               & $\sigma_{\omega}$& $0.2$\,rad/s      &  \\
Gravity (per-axis)                  & $\sigma_{g}$     & $0.05$\,g         &  \\ \midrule
\multicolumn{4}{l}{\textbf{Odom update-rate randomization}} \\ \hline
Update frequency (steps)            & $n_{\text{odom}}$ & $\mathcal{U}\{2,6\}$ & Odometry observation is held constant for $n$ env-steps to mimic drop-outs. \\ \midrule
\multicolumn{4}{l}{\textbf{Height-map sensing noise}} \\ \hline
White noise (cells)                 & $\sigma_{h}$     & $0.02$\,m         & Gaussian sampled per-cell and per-step. \\
Offset noise (cells)                & $\Delta_h$       & $0.02$\,m         & Height-bias (applied to all cells, Gaussian sampled per episode). \\
Roll / Pitch noise                  & $\sigma_{\phi},\sigma_{\theta}$ & $0.04$\,rad & Projection frame tilt, Gaussian per-env. \\
Yaw noise                           & $\sigma_{\psi}$  & $0.08$\,rad       & Map frame rotation, Gaussian per-env. \\
Sensor delay                        & $d_{\max}$       & $\mathcal{U}\{0,3\}$     & Frames of latency before height-map update. \\
Update frequency                    & $n_{\text{map}}$ & $\mathcal{U}\{1,5\}$ & Heightmap refresh period (steps). \\
Bad-distance probability            & $p_{\text{bad}}$ & $0.01$            & Chance a cell is replaced by a random value. \\ \bottomrule
\end{tabular}}
\vspace{0.5em}
\caption{\textbf{Domain-randomization and noise settings used during training.} ``Offset" terms are fixed per episode; “white” terms are resampled each simulation step. Update-frequency variables simulate low-rate sensors by holding observations constant for a random number of steps.}
\label{table:domain-randomisation}
\end{table}

\subsection{Training Data Distribution}
\label{sec:data_distribbution}
After the MPT stage, when training our pipelines on our own data, we train on 123 clips of human motion data collected from our pipeline. We also include during the training process the 10 clips of flat terrain walking data from LaFan ~\cite{harvey2020robust} that we used during MPT. Since we sample to be class balanced per-clip, this means we train ~90\% on our own data and 10\% on LaFan data. 
The motion distribution of our 123 collected clips is reported in Table \ref{table:clip-distribution}.

\begin{table}[ht]
\centering
\resizebox{1\linewidth}{!}{%
\begin{tabular}{@{}lcl@{}}
\toprule
\rowcolor[HTML]{E6E6E6}
\textbf{Category} & \textbf{Keyword examples} & \textbf{\#\,Sequences} \\ \midrule
Climbing up stairs        & ``climb up'', ``walk up stairs (backwards)'', ``step‑up platform / block''      & 39 \\
Climbing down stairs      & ``climb / walk down stairs'', ``down stairs backwards / sideways''  & 36 \\
Both up \& down stairs    & ``up then down'', ``mixed up / down'', ``two people opposite directions''& 13 \\
Standing (only)           & ``standing up''                                                     & 7  \\
Sitting (only)            & ``sit down'', ``sitting on sofa / chair / bench''                   & 12 \\
Both sitting \& standing  & ``sit then stand''                                                  & 5 \\
Terrain traversing        & ``stepping stones / chairs'', ``side‑walk'', ``step on board / block''  & 11 \\ \midrule
\textbf{Total}            &                                                                     & \textbf{123} \\ \bottomrule
\end{tabular}}
\vspace{0.5em}
\caption{\textbf{Categorization and distribution of our 123 self-collected training clips.}}
\label{table:clip-distribution}
\end{table}

\section{Robot Deployment}
\label{sec:appendix:robot-deployment}

We implement deployment code in C++ using ROS and the Unitree SDK 2 to enable fast running on the onboard Jetson Orin NX at 50Hz. We use gains of $Kp=75$ and $Kd=2$ for all joints except the ankle, where we use $Kp=20$ and $Kd=0.1$ and $Kd=0.2$ for roll and pitch, respectively, following \cite{zakka2025mujocoplayground}.

\subsection{Stages of Deployment in Real}

\begin{figure*}[t]
  \centering
  \begin{subfigure}{0.48\linewidth}
    \includegraphics[width=\linewidth,clip,trim=0cm 0cm 0cm 0cm]{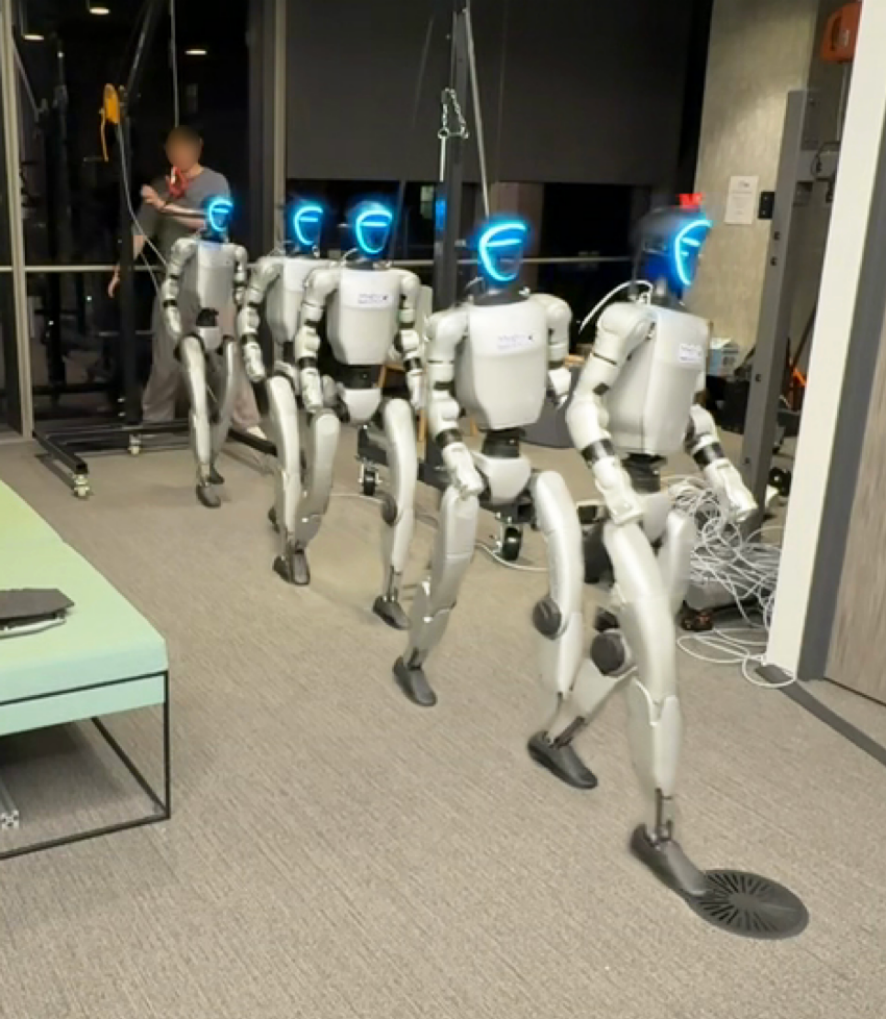}
    \label{fig:real_eval_a}
  \end{subfigure}\hfill
  \begin{subfigure}{0.51\linewidth}
    \includegraphics[width=\linewidth,clip,trim=0cm 0cm 0cm 0cm]{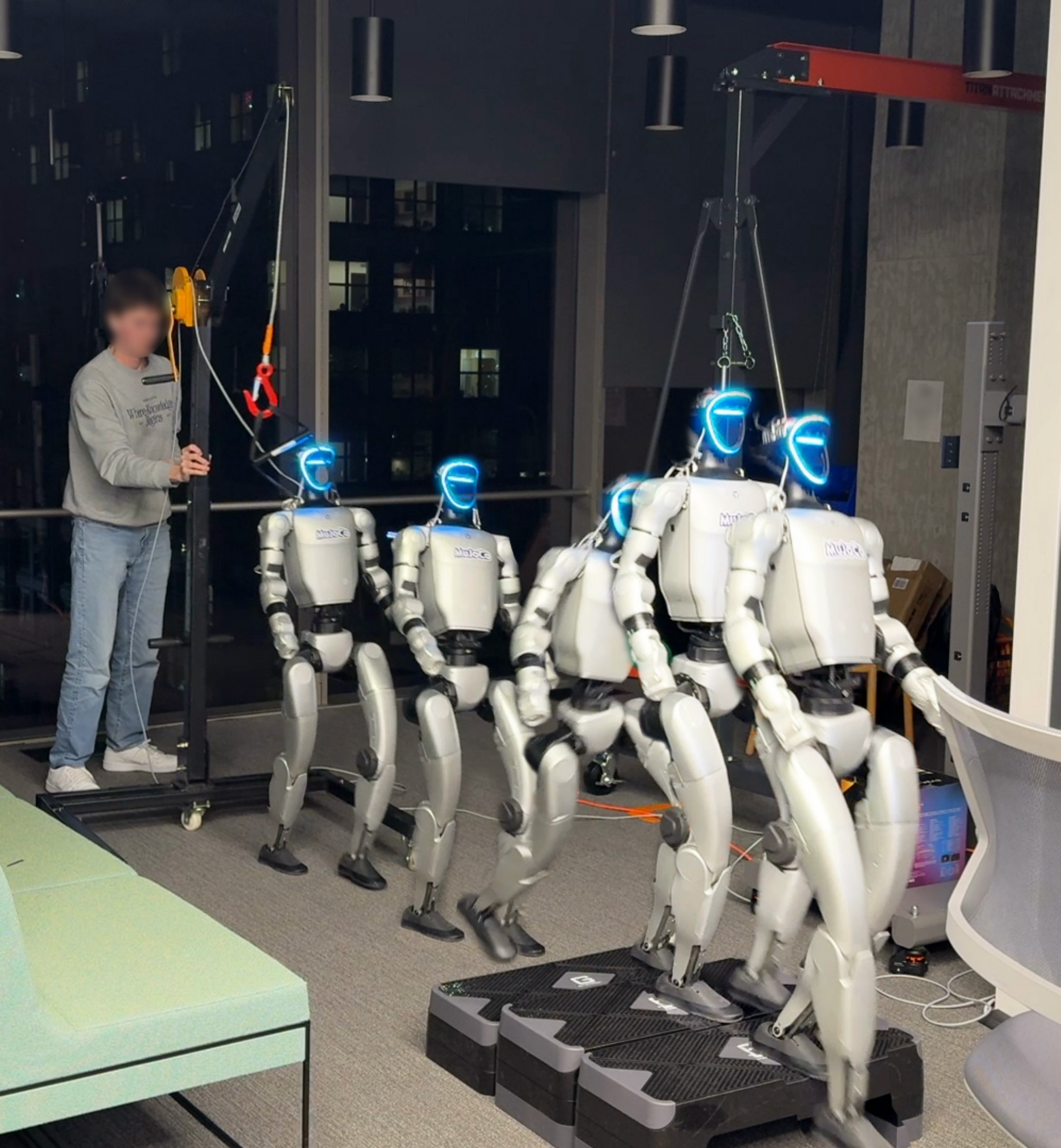}
    \label{fig:real_eval_c}
  \end{subfigure}

  \vspace{-0.5em}
  \caption{\textbf{Evaluation of policies at multiple stages.} Due to the multi-stage nature of our pipeline, we evaluate the performance on real at multiple stages before going out into the real world. \textit{Left}: trying our MoCap Pre-Trained policy. \textit{Right:} Trying the first generalist in a lab environment tracking a pre-defined root trajectory.}
  \label{fig:progressive-deployment}
  \vspace{-1em}
\end{figure*}

 We employed a progressive evaluation approach to deploying our policies, to gradually debug capabilities in the real world. The first stage was being able to track motion capture data from MPT policies. We then distilled motion capture policies trained on a large dataset to test various approaches to distillation; this is how we found that root position conditioning works better than root velocity. Finally, we started to deploy heightmap conditioned distilled policies from our full pipeline. Figure \ref{fig:progressive-deployment} shows two examples of this.

\end{document}